%% file: neurips_2025.tex
\documentclass{article}

\usepackage[final,dandb]{neurips_2025}

\usepackage[utf8]{inputenc} 
\usepackage[T1]{fontenc}    
\usepackage{hyperref}       
\usepackage{url}            
\usepackage{booktabs}       
\usepackage{arydshln}
\usepackage{ragged2e}
\usepackage{amsfonts}       
\usepackage{graphicx}       
\usepackage{nicefrac}       
\usepackage{microtype}      
\usepackage{xcolor}         
\usepackage{array}
\usepackage[table]{xcolor}

\usepackage{multirow}
\newcommand{\shline}{\specialrule{0.8pt}{0pt}{0pt}}
\usepackage{subcaption}
\usepackage{stfloats}
\usepackage{amsmath}
\usepackage[linesnumbered,ruled,vlined]{algorithm2e}
\usepackage{wrapfig}

\definecolor{myblue}{HTML}{2E54A1} 
\definecolor{mygreen}{HTML}{005826} 
\usepackage{wrapfig}

\usepackage{xspace}
\makeatletter
\DeclareRobustCommand\onedot{\futurelet\@let@token\@onedot}
\def\@onedot{\ifx\@let@token.\else.\null\fi\xspace}
\def\eg{\emph{e.g}\onedot}

\def\etc{\emph{etc}\onedot}

\makeatother

\title{PhyBlock: A Progressive Benchmark for Physical Understanding and Planning via 3D Block Assembly
}

\author{%
Liang Ma\textsuperscript{1}$^{*}$,~~Jiajun Wen\textsuperscript{2}\footnotemark[1],~~Min Lin\textsuperscript{3}\footnotemark[1],~~Rongtao Xu\textsuperscript{1,4}\footnotemark[1],~~\textbf{Xiwen Liang}\textsuperscript{3}\footnotemark[1],~~\textbf{Bingqian Lin}\textsuperscript{5},
\\\textbf{Jun Ma}\textsuperscript{3},~~\textbf{Yongxin Wang}\textsuperscript{1},~~\textbf{Ziming Wei}\textsuperscript{3},~~\textbf{Haokun Lin}\textsuperscript{1},~~\textbf{Mingfei Han}\textsuperscript{1},~~\textbf{Meng Cao}\textsuperscript{1},\\
~~\textbf{Bokui Chen}\textsuperscript{2}$^{\dagger}$,~~\textbf{Ivan Laptev}\textsuperscript{1},~~\textbf{Xiaodan Liang}\textsuperscript{1,3}$^{\dagger}$ \\
 \textsuperscript{1}Mohamed bin Zayed University of Artificial Intelligence\\\textsuperscript{2}Tsinghua Shenzhen International Graduate School, Tsinghua University, China\\\textsuperscript{3}Shenzhen Campus of Sun Yat-sen University~~\textsuperscript{4}Spatialtemporal AI~~\textsuperscript{5}Shanghai Jiao Tong University \\
 {\small{\textsuperscript{*}Authors contributed equally to this research.~~\textsuperscript{\dag}Corresponding author.}}\\
 \textbf{\url{https://phyblock.github.io/}}%
}

\begin{document}

\maketitle

\begin{abstract}
While vision-language models (VLMs) have demonstrated promising capabilities in reasoning and planning for embodied agents, their ability to comprehend physical phenomena, particularly within structured 3D environments, remains severely limited. To close this gap, we introduce PhyBlock, a progressive benchmark designed to assess VLMs on physical understanding and planning through robotic 3D block assembly tasks. PhyBlock integrates a novel four-level cognitive hierarchy assembly task alongside targeted Visual Question Answering (VQA) samples, collectively aimed at evaluating progressive spatial reasoning and fundamental physical comprehension, including object properties, spatial relationships, and holistic scene understanding. PhyBlock includes 2600 block tasks (400 assembly tasks, 2200 VQA tasks) and evaluates models across three key dimensions: partial completion, failure diagnosis, and planning robustness. We benchmark 23 state-of-the-art VLMs, highlighting their strengths and limitations in physically grounded, multi-step planning. 
Our empirical findings indicate that the performance of VLMs exhibits pronounced limitations in high-level planning and reasoning capabilities, leading to a notable decline in performance for the growing complexity of the tasks.
Error analysis reveals persistent difficulties in spatial orientation and dependency reasoning. 
We position PhyBlock as a unified testbed to advance embodied reasoning, bridging vision-language understanding and real-world physical problem-solving.




\end{abstract}

\section{Introduction}
\label{sec:intro}

\begin{figure}[t]
  \centering
   \includegraphics[width=0.98\linewidth]{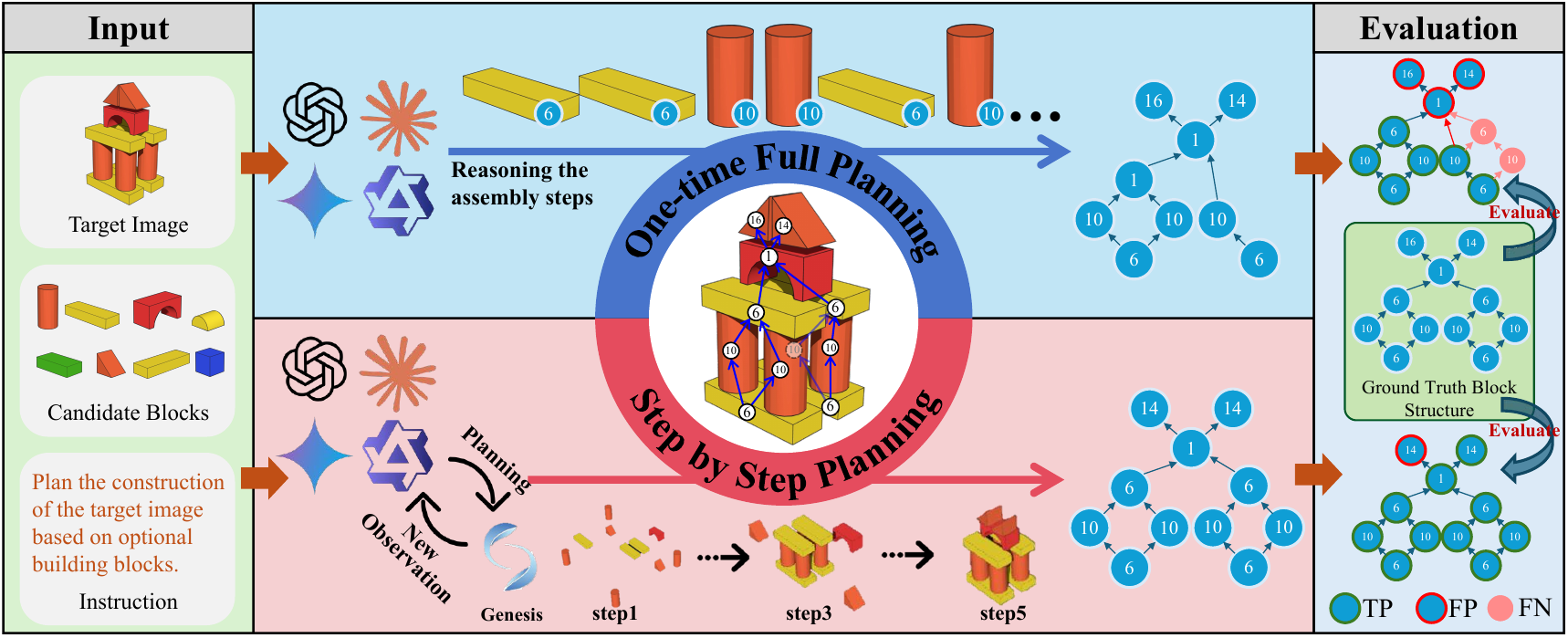}
   \vspace{-1mm}
   \caption{Assembly Planning Task in \textbf{PhyBlock}. Here shows inference setting of two planning strategies(one-time full planning and step-by-step planning).}
   \label{fig:inference_setting}
   \vspace{-2mm}
\end{figure}

Understanding physical interactions and spatial relationships is crucial for embodied agents tasked with manipulating and navigating complex real-world environments. Recent Vision–Language Models (VLMs), such as GPT-4o \cite{openai_hello_gpt_4o}, Claude-3.7 \cite{anthropic_claude_3_7_sonnet}, and Gemini 2.0 \cite{google_gemini_ai_update_2024}, have made impressive strides in multimodal reasoning, yet their grasp of physical-world characteristics—such as object stability, spatial support, and realistic multi-step planning—remains limited. As illustrated in Figure~\ref{fig:inference_setting}, 3D block assembly tasks serve as an intuitive testbed for these capabilities, encapsulating fundamental physical concepts like gravity (e.g., stability of constructed blocks), structural dependencies (e.g., correct block structure should be determined based on the desired target image), and geometric constraints. Accurately evaluating whether VLMs internalize such physical priors is critical, especially when they serve as high-level planners in hierarchical agent systems (e.g., System 2 in GR00T-N1~\cite{bjorck2025gr00t} and Helix~\cite{figure2025helix}). These systems rely on physical awareness to generate actionable plans for low-level controllers (System 1), bridging abstract reasoning with real-world execution. 

Existing benchmarks \cite{james2020rlbench, liu2023libero, shridhar2020alfred, feng2025reflective, zhang2024vlabench} still suffer from two critical limitations\cite{xiang2025aligning}: 1) Perception dominance without planning capability. Current frameworks primarily emphasize perceptual understanding while neglecting long-horizon planning, resulting in models that excel at single-step reasoning but demonstrate inadequate inference capacity in complex scenarios; 2) Unrealistic physical assumptions. The prevailing assumption of objects existing in idealized states while ignoring real-world physical interactions significantly undermines their practical applicability. Consequently, we lack a rigorous yardstick that couples high-level language reasoning with the dynamic constraints of the physical world, leaving open whether modern VLMs truly understand how objects interact in three dimensions.

To benchmark physical understanding and planning capability, we adopt interactive 3D blocks, as they intuitively embody fundamental physical concepts, such as stability, support, and spatial relationships, in a clear and interpretable manner. Leveraging a physics-based simulator, we construct realistic 3D scenes that dynamically respond to interactions, enabling systematic evaluations of increasingly complex, multi-step tasks.

\begin{figure*}[t]
  \centering
   \includegraphics[width=0.98\linewidth]{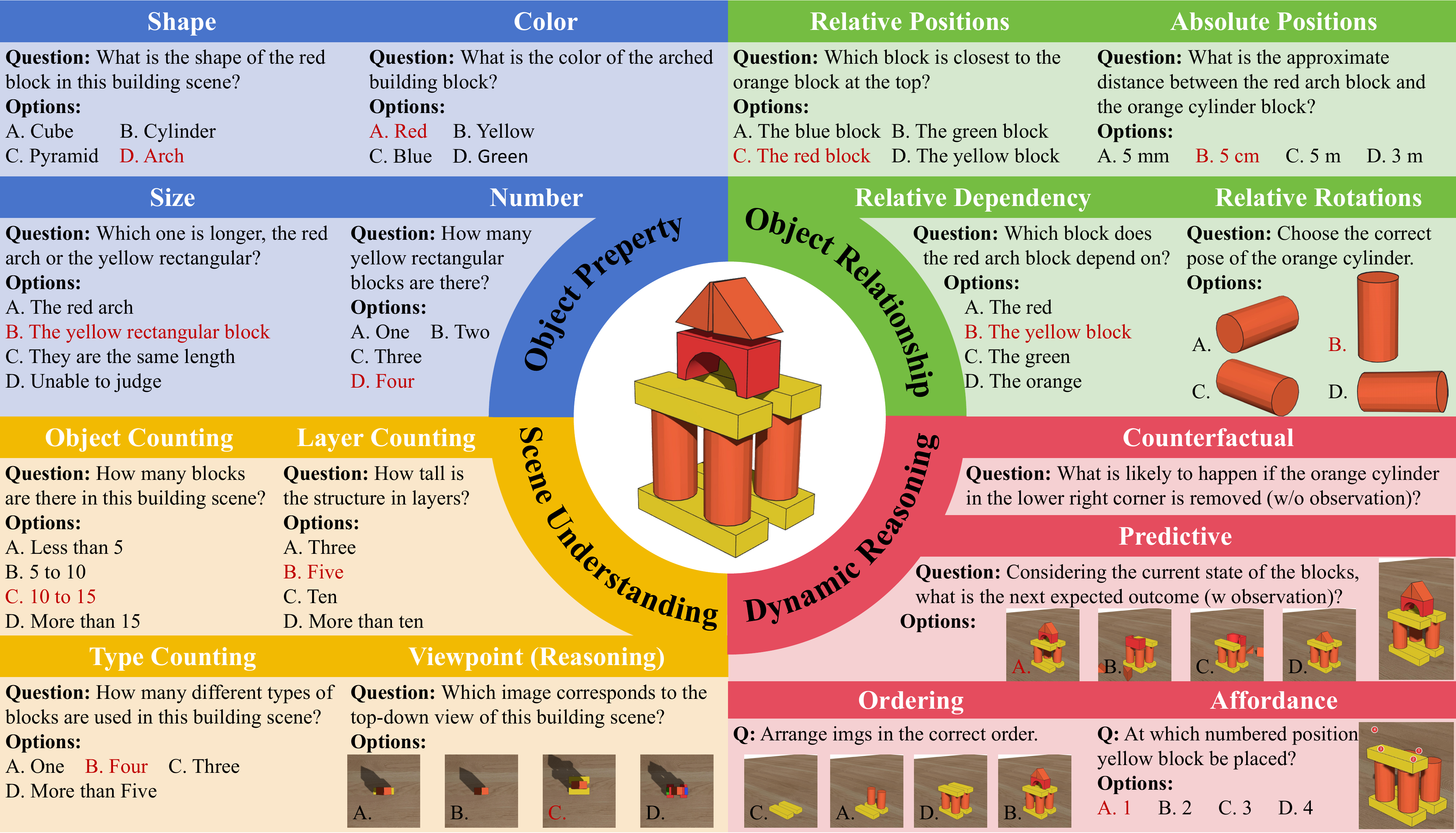}
   \vspace{-1mm}
   \caption{Physics Understanding VQA in \textbf{PhyBlock}. We construct a compact set of questions per 3D assembly scene, covering four key dimensions of physical and spatial reasoning to assess diverse aspects of the model’s understanding of 3D block assembly.}
   \label{fig:PHYPART}
   \vspace{-5mm}
\end{figure*}

Building on this insight, we present \textbf{PhyBlock}, a comprehensive two-branch benchmark explicitly designed to assess the physical reasoning capabilities of modern VLMs.
The first branch, \textbf{Hierarchical Assembly Planning} (shown in Figure~\ref{fig:inference_setting}), evaluates model’s capacity to plan and reason about spatial arrangements through step-by-step interactions in a physics-aware simulator.  This planning branch features 400 systematically constructed scenes across four ascending difficulty tiers (Basic, Simple Combinations, Complex Structures, and Advanced Spatial Planning), culminating in assemblies that involve up to 22 distinct blocks. 
The second branch, \textbf{Physical-Understanding VQA} (shown in Figure~\ref{fig:PHYPART}), measures model’s explicit understanding of physical concepts.
The VQA branch comprises 2,200 rigorously curated questions spanning 16 semantic categories including object attributes, relational reasoning, scene dynamics, and counterfactual inference.
Drawing inspiration from cognitive-development research, particularly the observation that structured block play enables children to internalize complex spatial and physical principles, we model eight LEGO-like block geometries in five distinct colors within the Genesis physics simulator, ensuring uniform and physically plausible interactions. This design not only captures key real-world regularities but also leaves headroom for future extensions that integrate low-level motor actions and control policies, thereby bringing the benchmark even closer to embodied deployment scenarios.
To further guarantee dataset quality and rigor in evaluation, we encode essential dependencies and spatial relationships between blocks with an Activity-on-Vertex (AOV) graphs (detailed in Section \ref{subsec:3.2}), and construct manually verified Visual Question Answering (VQA) tasks through a robust, multi-stage process combining automated generation and rigorous human validation (detailed in Section~\ref{subsec:3.3}). This careful design supports clear diagnostics, precise scoring, and reproducible analysis.

Building on PhyBlock, we conduct a comprehensive evaluation of 23 state-of-the-art open-source and closed-source vision-language models (VLMs) \cite{achiam2023gpt,openai2023gpt4v,bai2023qwen,yang2024qwen2, li2024llava, lin2023video}, covering diverse architectures and scales. 
Empirical results uncover three consistent trends.
First, a steep \emph{performance cliff}: mean planning $F_{1}$ scores drop by more than half from the simplest to the most challenging assembly tier, and no model maintains high recall once tasks demand long-horizon sequencing or hidden-support reasoning.  
Second, a \emph{perception–reasoning gap}: models answer low-level questions about colour or shape with high accuracy, yet accuracy collapses on counterfactual, causal, or affordance queries, mirroring the assembly failures that stem from unmodelled dynamics.  
Third, a pair of \emph{universal error modes}: (i) mis-estimated the orientation of blocks that lead to systematically incorrect block poses, and (ii) ignored support dependencies that violate basic stability—together accounting for the majority of mistakes across architectures.  Notably, enabling thinking mode prompting in larger models leaves these two error modes virtually unchanged, indicating that more text tokens alone cannot compensate for missing physical priors.
Collectively, these findings expose a fundamental shortfall in current multimodal pre-training: while today’s VLMs perceive objects well, they still lack the physical insight and sequential reasoning needed for reliable embodied planning.  Bridging this gap will require architectures and training regimes that fuse rich visual embeddings with explicit physics reasoning and interactive feedback, charting a path toward truly capable embodied agents.


Our contributions are as follows:
\begin{itemize}
    \item \textbf{PhyBlock benchmark.} We present a unified testbed for physical understanding and multi-step planning, built on interactive 3-D blocks in a high-fidelity physics simulator with strict guarantees on spatial precision and physical feasibility.
    \item \textbf{Progressive dataset.} We release a cognitively inspired dataset of 3D scenes, dependency graphs, step-wise plans, and 2,200 validated VQA pairs that scale smoothly from simple stacks to 22-block assemblies.
    \item \textbf{Comprehensive evaluation.} We assess 21 leading VLMs (13 proprietary, 8 open-source) and show that, despite strong perceptual skills, all models falter on complex spatial reasoning, physical inference, and long-horizon planning—exposing key challenges for future embodied intelligence.

\end{itemize}

\section{Related Work}

\subsection{Benchmarks and Datasets}
Current benchmarks for evaluating physical understanding in embodied AI systems exhibit critical limitations. While manipulation-focused benchmarks like RLBench and LIBERO \cite{james2020rlbench,li2024evaluating,liu2023libero} assess basic object interaction skills, they fail to systematically evaluate fundamental physical comprehension including object stability, structural integrity, and spatial relationships. Navigation-oriented benchmarks \cite{shridhar2020alfred,szot2021habitat} emphasize environment interaction but neglect the precise physical reasoning required for structured assembly tasks. Most existing datasets either lack physical grounding \cite{brown2020language} or suffer from limited scalability in 3D construction scenarios \cite{nasiriany2024robocasa,walke2023bridgedata}. 
Recent efforts like Reflective Planning \cite{feng2025reflective} and VLABench \cite{zhang2024vlabench} advance multi-stage reasoning but neglect physical constraints (e.g., gravity, structural dependencies) critical for assembly.

PhyBlock fills these gaps as the first benchmark evaluating 3D block assembly. It combines VQA and planning tasks to assess physical reasoning (stability, object properties) and multi-step planning, bridging perception and real-world understanding.

\subsection{Vision-Language Models}

Recent advances in VLMs have expanded multimodal reasoning capabilities and enabled their successful application across diverse domains\cite{han2025roomtour3d,hao2025conav,liang2023cornav,lin2025navcot,lin2025evolvenav,cao2025ground,lin2025inframind,cao2025video}, yet their understanding of physical phenomena remains limited. While foundational models like GPT-4 \cite{achiam2023gpt} and GPT-4V \cite{openai2023gpt4v} demonstrate strong visual-language alignment, and open-source alternatives \cite{vicuna2023,touvron2023llama} enable broader experimentation, these systems show fundamental gaps in physical comprehension. Specialized architectures like region-level VLMs \cite{cai2024vip,chen2023shikra,peng2023kosmos} improve spatial awareness for object localization, while video-based models \cite{lin2023video,tan2024koala,weng2024longvlm,cao2022locvtp} enhance temporal reasoning, yet neither adequately addresses the physical understanding required for structured 3D assembly. Current benchmarks \cite{cai2024temporalbench,fu2024video} primarily assess static perception or functional affordances rather than the core physical reasoning capabilities needed for tasks involving object properties, structural stability, and spatial dependencies. 

This limitation persists despite architectural innovations in models like LLaVA \cite{liu2023visual,zhang2024llavanext-video} and MiniGPT-4 \cite{zhu2023minigpt}, which integrate vision encoders with language models but lack explicit representations of physical constraints. Our analysis reveals that existing VLMs struggle with the intuitive physical reasoning that humans employ when manipulating objects - particularly in understanding how spatial relationships affect structural integrity, predicting physical outcomes of actions, and maintaining consistency across multi-step assembly sequences. PhyBlock addresses these gaps by providing the first systematic evaluation framework specifically designed to probe models' physical understanding through the lens of 3D block assembly, complementing existing benchmarks that focus on perception or functional reasoning.

\subsection{Physical Understanding}

Physical understanding \cite{bobrow1984qualitative,hespos2016five,mccloskey1983intuitive} has widespread applications spanning visual reasoning \cite{lerer2016learning,yiclevrer,cao2024physgame}, embodied AI \cite{agrawal2016learning,byravan2017se3}, \etc. The primary works \cite{battaglia2016interaction,lerer2016learning,yiclevrer,wang2025wisa}focused on simple scenarios where visual primitives (\eg, spheres, cubes) are restricted to a limited set of interactions. The follow-up works \cite{chow2025physbench,zhao2025mmvu,wang2024compositional} extend the scope to more realistic scenes with real-world objects and complex backgrounds. Wang \emph{et.al} \cite{wang2024compositional} introduces the 4D scene representation to simultaneously model dynamic properties of objects and multi-object interactions. PhysBench \cite{chow2025physbench} is introduced to evaluate VLMs’ physical world understanding capability across a more diverse and comprehensive tasks. Another stream of works focus on spatial intelligence, which requires the understanding the objects positions in 3D space and their relationships in-between. 

Compared with existing benchmarks, our proposed PhyBlock offers several key advantages in evaluating physical understanding and planning. First, unlike prior datasets that are either limited to passive visual prediction or constrained to toy-like synthetic scenes, PhyBlock introduces a goal-conditioned block stacking task grounded in high-fidelity physics simulation. Second, PhyBlock supports interactive and constructive physical reasoning. Instead of merely recognizing or forecasting physical events, VLMs are required to plan and generate a sequence of physically plausible actions to achieve a specified structural goal, which aligns more closely with real-world embodied scenarios.

\section{PhyBlock}



In human cognitive and educational psychology, structured block play has been hypothesized to cultivate essential cognitive skills, including estimation, measurement, pattern recognition, part–whole relations, visualization, symmetry, transformation, and balance \cite{verdine2014deconstructing}. Given the cognitive benefits of structured block play, we extend this framework to evaluate embodied agents in three fundamental capabilities: visual alignment and pose estimation, spatial reasoning, and long-horizon planning.

In this section, we introduce the hierarchical capability levels in Sec. \ref{subsec:3.1}. Then in Sec. \ref{subsec:3.2}, we detail the capability-oriented data collection process. Next, we present the construction of the Physics Understanding VQA dataset in Sec. \ref{subsec:3.3}, which enables fine-grained evaluation of scene perception and physical reasoning. Finally, the overall dataset construction process is demonstrated in Sec. \ref{subsec:3.4}.

\subsection{Hierarchical Capability Levels} \label{subsec:3.1}

In human cognitive and educational psychology, structured block play has been hypothesized to cultivate essential cognitive skills. Following human cognitive skill, we propose \textbf{PhyBlock} to benchmark VLMs on Robotic 3D Block \textbf{Assembly} \textbf{Planing} task. 

To systematically assess these core capabilities, we construct a hierarchical capability levels for embodied  3D Block Assembly inspired by the developmental stages of children's cognitive growth. Specifically, PhyBlock is curated in a hierarchical capability levels including basic perception, simple combinations, complex structures, and advanced spatial planning.





\noindent
\textbf{Level-1 Basic Perception.} The model is required to identify and select correct blocks from a component library based on a reference diagram. Tasks involve up to four blocks with minimal variation in type and color, focusing on visual feature recognition and matching accuracy.



\noindent
\textbf{Level-2 Basic Simple Combinations.}
Building on Level-1, this stage evaluates elementary structural reasoning. The model must select fewer than six relevant blocks and generate a valid assembly sequence with up to three vertical layers, respecting basic support relations and spatial dependencies.

\noindent
\textbf{Level-3 Complex Structures.}
At this level, the model must not only select the necessary blocks but also plan an optimal assembly sequence, ensuring a logical and stable construction process. Compared to Level-2, the dependency relationships are significantly more complex. The scenarios in this level contain up to 12 blocks with a maximum of 8 layers, demanding advanced 3D spatial reasoning and multi-step decision-making capabilities.

\noindent
\textbf{Level-4 Advanced Spatial Planning.}
As the highest complexity level, this stage requires the model to execute systematic planning for assembling large-scale structures under complex spatial constraints. The scenarios involve up to 22 components, challenging the model’s ability to develop a global understanding of intricate 3D structures and execute long-horizon spatial reasoning and planning.



\subsection{AOV-Based Assembly Evaluation} \label{subsec:3.2}

Block assembly exhibits non-Markovian dependencies: placing a block on an upper layer requires proper support from lower layers. While inter-layer construction must follow strict temporal order, within-layer actions can often be executed in parallel. Final-state-only evaluation fails to disentangle errors in physical reasoning, planning, and control. 
To better analyze the hierarchical and sequential constraints inherent in physical assembly, we introduce the Activity-on-Vertex (AOV) network, which models blocks as vertices and their assembly dependencies as directed edges, as illustrated in Figure~\ref{fig:inference_setting}. This graph-based representation captures both inter-layer temporal dependencies and intra-layer parallelism, enabling fine-grained analysis of planning behaviors.



This dual-representation enables more rigorous evaluation by: (1) computing intermediate metrics to quantify partial completion, even in failure cases; and (2) diagnosing failure modes via systematic analysis of dependency violations, such as missing prerequisites or conflicting operations.

This AOV framework enables: (1) fine-grained assessment via intermediate completion metrics, (2) diagnostic analysis of failure modes based on violated dependencies, and (3) evaluation of planning robustness across valid sequence variations.
Details of the AOV-based evaluation algorithm are provided in \textcolor{blue}{Appendix~\ref{sub:aov}}.


\subsection{Physics Understanding VQA}
\label{subsec:3.3}
To evaluate an agent's capability for physical reasoning, we propose a comprehensive set of questions targeting diverse aspects of 3D block assembly understanding. As illustrated in Figure~\ref{fig:PHYPART}, the questions are grouped into four major categories: \textbf{Object Property}, \textbf{Object Relationship}, \textbf{Scene Understanding} and \textbf{Dynamic Reasoning}, focusing on both static perception and dynamic physical reasoning. Details for each are provided below.

\textbf{Category 1: Object Property}. 1) Shape: Identify the geometric shape of a given block. 2) Color: Determine the color of a specified object. 3) Size: Compare the dimensions (e.g., length or height) of two blocks. 4) Number: Count how many blocks of a particular color or type are present. This category assesses the agent's ability to understand basic attributes of individual objects.

\textbf{Category 2: Object Relationship}. 1) Relative Positions: Analyze the relative positional relationships between blocks, such as proximity and distance. 2) Absolute Positions: Estimate the spatial relationships between blocks by providing concrete numerical values with physical units. 3) Relative Dependency: Identify which blocks depend on or support others. 4) Relative Rotations: Determine the relative rotational relationships between blocks. This category focuses on spatial and logical relationships among multiple blocks.

\textbf{Category 3: Scene Understanding}. 1) Object Counting: Estimate the number of blocks present in the scene. 2) Layer Counting: Infer how many vertical layers the construction consists of. 3) Type Counting: Count the number of distinct block types (e.g., cube, arch, cylinder). 4) Viewpoint: Match a given single-view image with its corresponding 3D scene configuration. This category assesses the agent's holistic perception of the environment, requiring recognition of object presence, spatial composition, and view-consistent scene interpretation.

\textbf{Category 4: Dynamic Reasoning}. 1) Counterfactual: Predict what will happen if a supporting block is removed. 2) Predictive: Anticipate the next step or possible continuation of the current assembly. 3) Ordering: Determine the correct temporal or structural sequence of subassemblies. 4) Affordance:  Decide where a given block can be stably placed. This category evaluates the agent's understanding of physical dynamics, structural stability, and potential consequences of actions

These question types collectively establish a progressive and fine-grained benchmark for evaluating physical understanding in both VLMs and embodied agents. PhyBlock emphasizes grounded reasoning beyond visual recognition, targeting real-world generalization and planning competence.

\subsection{Dataset Construction} \label{subsec:3.4}
\textbf{Construction of Simulated Block Assets.}
We construct a parametric simulated block library inspired by global standards (e.g., LEGO®, Mega Bloks®), covering eight shapes and five colors. Detailed geometric specifications and texture mappings are provided in the \textcolor{blue}{Appendix~\ref{app:generate_assets}}.

\noindent
\textbf{Construction of Block Assembly Scenes.}
We use the Genesis simulator to construct scenes with precise control, recording each block’s pose and dependencies in structured JSON files for downstream analysis. 
Detailed procedures and examples are provided in the \textcolor{blue}{Appendix~\ref{app:generate_scenes}}.

\noindent
\textbf{Construction of Physics Understanding VQA.}
We introduce two data generation paradigms: LLM-based static VQA construction and simulation-driven dynamic VQA generation, targeting both perception and physically grounded reasoning tasks. 
Refer to \textcolor{blue}{Appendix~\ref{app:generate_VQA}} for details.

\noindent
\textbf{Data Augmentation.} 
The Level-4 block assembly scenes are generated by augmenting Level-3 scenes through compositional transformations. 
Due to the deliberate focus of PhyBlock as an \emph{evaluation benchmark} rather than a large-scale training corpus, we adopt a carefully balanced dataset size—comprising 400 assembly tasks and 2,200 VQA samples—which we found sufficient to reliably assess the physical reasoning and planning capabilities of modern VLMs without introducing redundant or overlapping samples. 
Each scene is generated within a high-fidelity physics simulator and manually verified to ensure physical validity and uniqueness, making large-scale expansion computationally expensive and conceptually unnecessary for diagnostic evaluation.

Importantly, our minimalist design—with simple geometric shapes and plain colors—reduces visual confounds and isolates core physical reasoning capabilities, enabling a controlled and cognitively interpretable evaluation. 
However, leveraging our open-source code interface, users can easily scale the dataset to millions of scenes through automated compositional transformations (e.g., color combinations, shape variations, scene compositions, and lighting adjustments) without compromising quality or physical realism.
This design philosophy balances rigorous benchmark fidelity with extensibility for future large-scale studies.

\section{Experiment}
\label{sec:exp}

\input{tables/main_results2}

\input{tables/sub_Reason}
\input{tables/vqa_results_new}




\begin{figure}[t]
    \centering
    \begin{subfigure}{6.75cm}
        \centering
        \includegraphics[width=\linewidth]{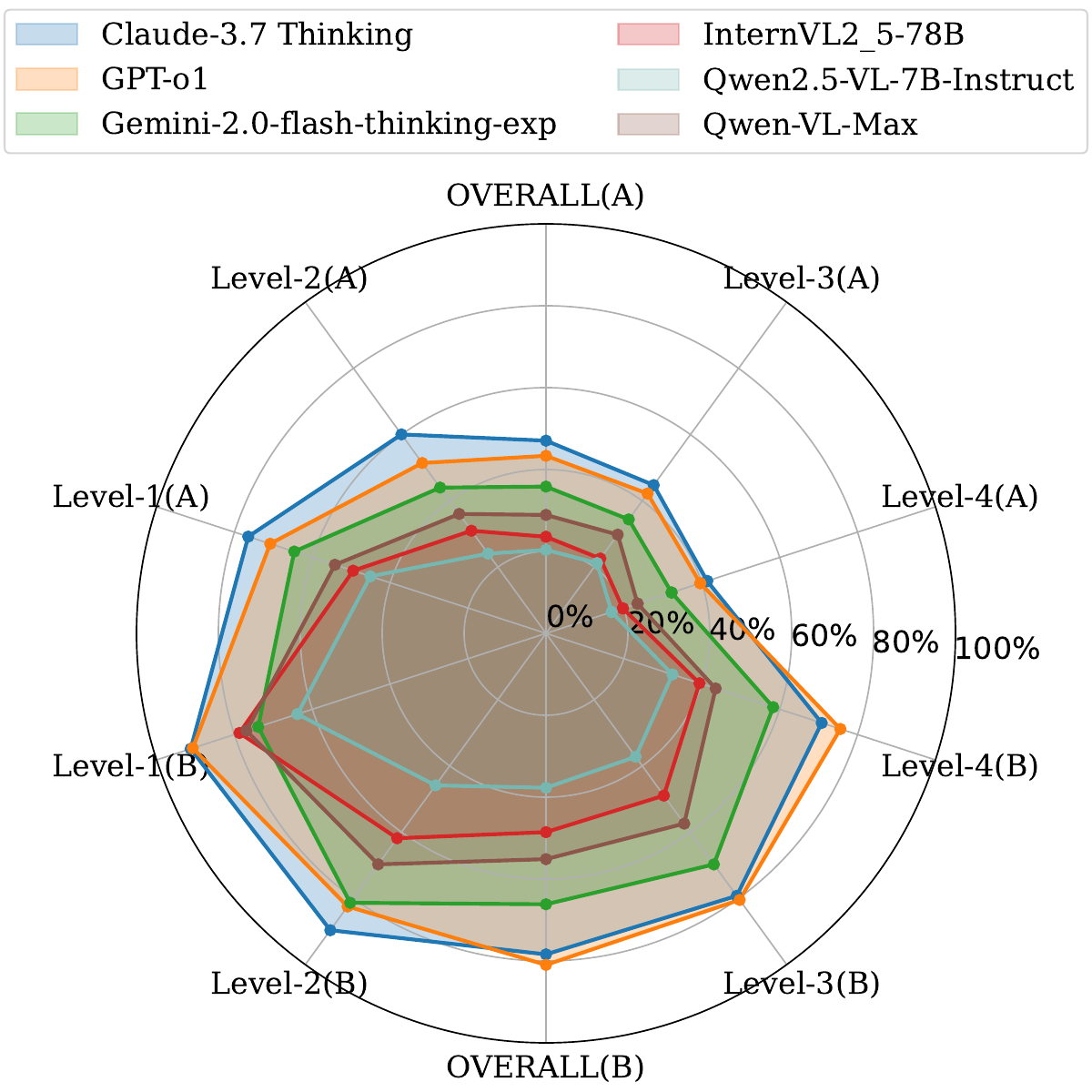}
        \caption{Task-level performance of mainstream models}
    \end{subfigure}
    \begin{subfigure}{6.25cm}
        \centering
        \begin{subfigure}{\linewidth}
            \centering
            \includegraphics[width=\linewidth]{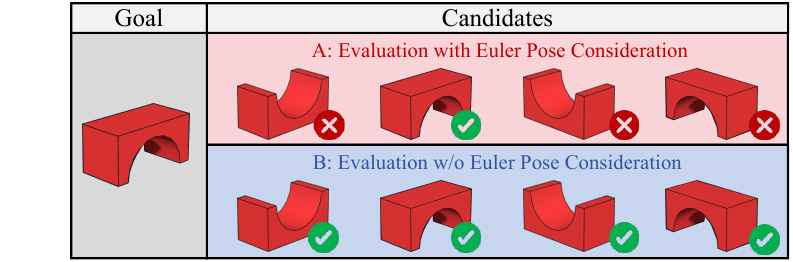}
            \caption{Two evaluation settings: \textcolor{red}{A} and \textcolor{blue}{B} }
        \end{subfigure}
        \begin{subfigure}{\linewidth}
            \centering
            \includegraphics[width=\linewidth]{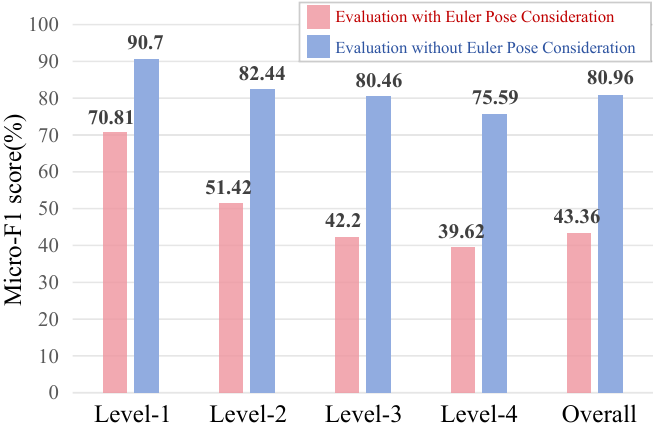} 
            \caption{Performance of \textbf{\textit{GPT-o1}} under two settings}
        \end{subfigure}
    \end{subfigure}

    \caption{Comprehensive Comparison of Mainstream Models Across Evaluation Dimensions. (a) We conduct a comprehensive comparison of six representative models under both \textcolor{red}{A} and \textcolor{blue}{B} evaluation settings across all four task difficulty levels. 
    (b) The differences between two Evaluation Setting are illustrated. For a detailed explanation, please refer to \textcolor{blue}{Appendix~\ref{sub:B2}}.
    (c) A focused analysis on GPT-o1 reveals its performance under the two evaluation settings. Interestingly, we observe a significant performance boost when the strict constraint on pose alignment is relaxed, highlighting the model's potential under less rigid spatial requirements.}
    \label{fig:Radar6Models}
    \vspace{-5mm}
\end{figure}


\begin{figure}[t]
  \centering
  \includegraphics[width=1.0\linewidth]{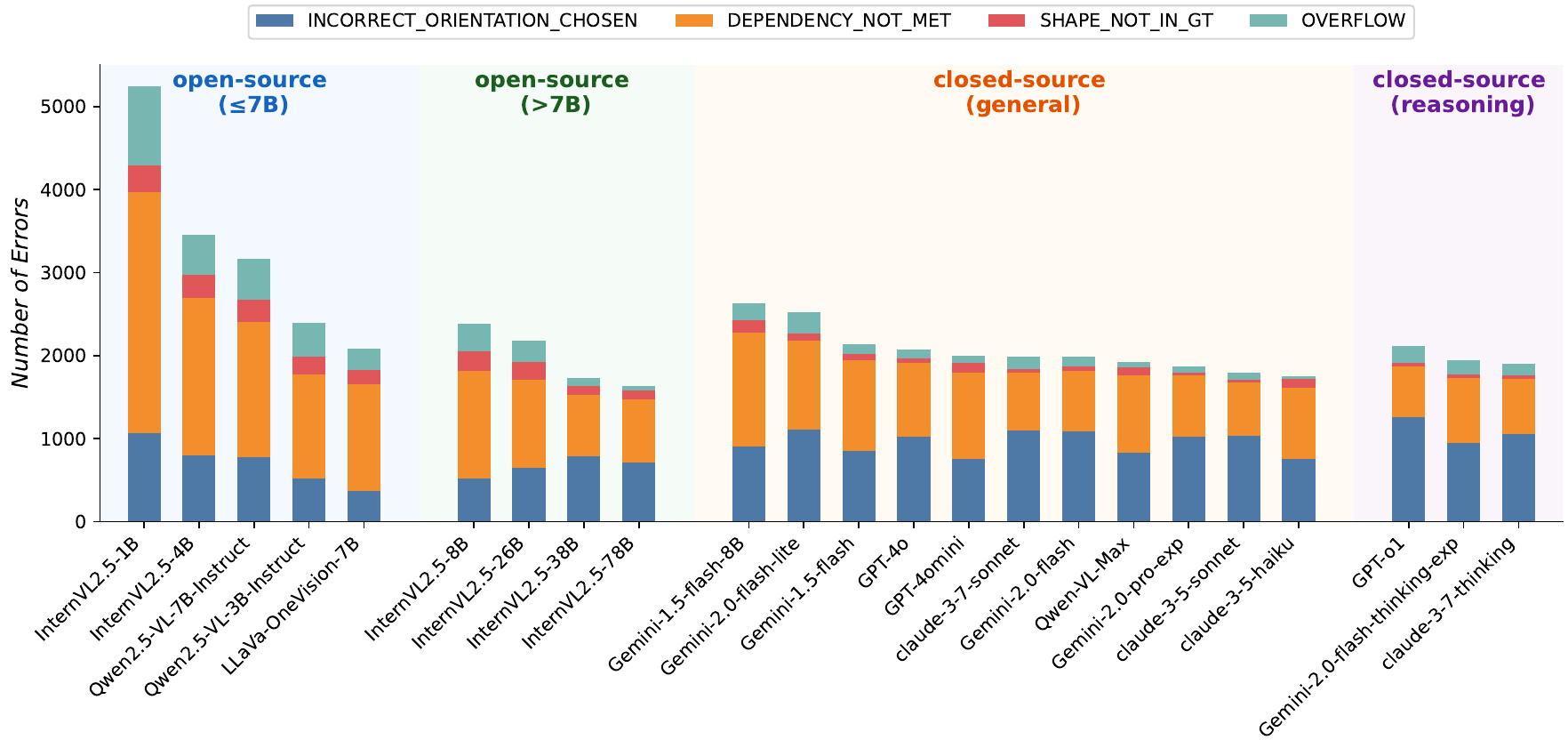}
    \vspace{-6mm}
   \caption{Error Type Analysis of Assembly Steps. We systematically analyze \textbf{four distinct types of errors} encountered during the planning process for each sample. A detailed definition and categorization of the four error types can be found in the \textcolor{blue}{Appendix~\ref{sub:errortype}}.}
   \vspace{-5mm}
   \label{Fig:Errors}
\end{figure}


\subsection{Setup}
\noindent\textbf{Evaluated Models:} We evaluate a range of state-of-the-art VLMs on the PhyBlock benchmark. Our evaluation includes fourteen proprietary models: GPT-O1 \cite{openai_o1}, GPT-4o \cite{openai_hello_gpt_4o}, GPT-4o-mini \cite{openai_gpt_4o_mini}, the Gemini-1.5 series \cite{team2024gemini}, the Gemini-2.0 series \cite{google_gemini_ai_update_2024}, Qwen-VL-Max \cite{bai2023qwen}, the Claude 3.5 series \cite{anthropic_claude_3_5_sonnet}, and the Claude 3.7 series \cite{anthropic_claude_3_7_sonnet}. Additionally, we assess eleven open-source models, including the LLaVA-OneVision series \cite{li2024llava}, the Qwen 2.5 series \cite{yang2024qwen2}, and the InternVL 2.5 series \cite{chen2024expanding}.

\noindent\textbf{Inference Setting:} In our inference setup, models are tasked with generating structured block assembly plans given a goal image, a set of candidate blocks, and a text instruction describing the construction objective. As shown in Figure \ref{fig:inference_setting}, we evaluate two distinct planning strategies:

\noindent\textbf{One-time Full Planning:} Given an initial scene (RGB-D scans), a goal image, and a textual instruction, the model generates a complete block assembly plan in a single forward pass—without iterative feedback or action history. This tests the model’s global planning ability.

\noindent\textbf{Step-by-Step Planning:} In an interactive simulator, the model receives step-wise visual observations and action histories, generating the next operation incrementally. This evaluates closed-loop planning grounded in environment feedback.

\noindent\textbf{Physics Understanding VQA Inference:} The model is queried with simple natural language questions based on block scene images from the Physics Understanding VQA dataset. Prompts target the question, testing the model’s intuitive physical reasoning via direct visual grounding.

\noindent\textbf{Evaluation Metrics:}
We evaluate performance using precision, recall, and F\textsubscript{1}-score, based on step-wise correctness defined by the AOV constraints: correct steps as True Positives (TP), incorrect as False Positives (FP), and missing required steps as False Negatives (FN). Micro-F\textsubscript{1} is computed across all samples and difficulty levels for overall performance.
In addition, to evaluate the agent’s physical perception and reasoning capabilities, we adopt a simple yet effective metric for the Physics Understanding VQA dataset. Since each question follows a multiple-choice format, we report the \textbf{accuracy}—the proportion of correctly answered questions—as the primary evaluation metric.

\paragraph{Random Baseline.}
To contextualize model performance, we include a random baseline that simulates an unskilled agent acting without perception or reasoning. 
For the 3D Block Assembly Planning task, actions are sampled uniformly at random from the valid block and placement space within the Genesis simulator. 
These random plans are scored using the same AOV-based precision, recall, and F$_1$ metrics, yielding an overall F$_1$ of about \textbf{8.8\%}, which represents the expected lower bound of purely stochastic assembly.
For the Physical Understanding VQA branch, answers are sampled uniformly from all options. 
The Ordering subtask requires ranking rather than single-choice selection, lead to slightly lower scores, but the overall accuracy remains around \textbf{24\%}.

\paragraph{Human Expert Upper Bound.}
To complement our quantitative evaluation of vision-language models, we additionally establish a human expert upper bound for reference. 
We conducted a controlled study involving \textbf{20 adult participants} with academic or engineering backgrounds related to embodied AI and robotics. 
Each participant was presented with \textbf{400 representative tasks} randomly sampled from our benchmark, covering both the assembly planning and physical understanding VQA branches. 
Participants were instructed to answer the questions or design task plans based on the same multimodal inputs (i.e., textual instructions and visual scenes) as used for model inference.

The collected responses were scored using the same evaluation metrics as the benchmark. 
On average, human experts achieved a \textbf{score of 378.83 out of 400}, corresponding to an overall accuracy of approximately \textbf{94.7\%}. 
This result provides an empirical upper bound for interpreting model performance and highlights the considerable gap that remains between current VLM capabilities and human-level reasoning in physically grounded planning and understanding tasks.


\subsection{Experiment Findings}

\textbf{Performance plummets with increasing task complexity.}

As shown in Table \ref{tab_main:main_results}, we evaluated a range of state-of-the-art vision-language models across four difficulty levels of vision-based block construction tasks. The results clearly demonstrate that current models struggle significantly as task complexity increases. The best-performing model, Claude 3.7 Sonnet, achieved the highest overall recall (47.15\%) and F\textsubscript{1} score (47.36\%), yet its performance still sharply declined from simpler (Level-1 recall 75.62\%, F\textsubscript{1} 76.78
\%) to more complex tasks (Level-4 recall 40.93\%, F\textsubscript{1} 41.82\%). This trend was consistently observed across all evaluated models, highlighting their limitations in handling complex spatial reasoning and multi-step planning tasks.

\textbf{Current models excel at object properties but remain challenged by complex physical inference.}
We report the evaluation results of various models on our proposed Physical VQA benchmark, as shown in Table \ref{tab_main:main_results_vqa}. Among the evaluated models, GPT-o3 achieves the highest overall accuracy (70.0\%), demonstrating strong generalization across diverse physical reasoning tasks. It notably excels in Object Property (e.g., 90.0\% in CO), Object Relationship (86.0\% in RD), and Scene Understanding (80.0\% in LC). Claude-3.5 Sonnet and Gemini-2.0-flash also show competitive performance, particularly in perceptual tasks such as Shape and Color, though their capabilities on reasoning-heavy tasks (e.g., Counterfactual and Affordance) remain more limited. 
These results highlight recent advances in multimodal models’ abilities to perceive and reason about physical properties, while also indicating that complex causal and temporal reasoning remains a challenging frontier.

\textbf{Incorrect Orientation Chosen dominate across models, highlighting universal spatial-reasoning gaps.}
Figure \ref{Fig:Errors} categorizes assembly errors into four types: orientation, dependency, shape, and overflow errors. Smaller models ($\leq$7B parameters, e.g., InternVL2.5-1B) show high error rates, especially in orientation and dependency tasks, while larger open-source models reduce errors but retain dependency issues. Commercial models (GPT-4o, Claude) outperform but still struggle with orientation chosen errors. Notably, reasoning-tuned models (GPT-o1, Claude-3.7-thinking) reduce dependency/overflow errors but not orientation chosen mistakes, underscoring spatial reasoning as a key challenge for future work. As an ablation study, we evaluate model performance under two evaluation paradigms and observe a significant performance boost when strict pose alignment constraints are relaxed, highlighting the model’s limitations in universal spatial reasoning (Figure~\ref{fig:Radar6Models}).
 Originally, we required models to directly predict absolute orientation or relative spatial coordinates, but this setting yielded near-zero success rates—even for tasks trivial to humans. After simplifying the formulation to high-level spatial reasoning, the tasks remained highly challenging.

\begin{wrapfigure}{r}{0.45\textwidth}
    \centering\small
    \captionof{table}{Results (\%) overview. Step-by-step interactive reasoning results with the environment.} 
    \label{tab_main:main_results}
    \setlength\extrarowheight{2pt}
    \normalsize
    \resizebox{0.45\textwidth}{!}{
    \begin{tabular}{l c c c}
        \toprule
        \multirow{2.5}{*}{\bf Model} &  
        \multicolumn{3}{c}{\bf \cellcolor[HTML]{CDD4DF} Overall Pref}\\

        \cmidrule(lr){2-4}& 
        \textit{$Prec$} & \textit{$Rec$} & \textit{$F_\textsubscript{1}$} \\
        
        \midrule
        GPT-4o & 90.1& 13.0 & 22.8\\
        Qwen2.5-VL-72B-Instruct & 94.9 & 35.9 & 52.1\\
        Qwen-VL-Max & 96.1 & 16.5 & 28.2\\
        Claude-3.7 Sonnet-Thinking & 69.7 & 23.9 & 35.6\\
        \bottomrule
    \end{tabular}}
\end{wrapfigure}

\textbf{Thinking Mode offers negligible benefit.}
Table \ref{tab_main:main_results} indicate that Claude 3.7 Sonnet performs nearly identically under normal inference and with thinking mode enabled, with a similar error distribution. This suggests that Thinking Mode reasoning provides little to no benefit on this benchmark. We posit that spatial understanding of block shapes relies more on the model’s intuitive processing rather than the generation of an extensive reasoning chain. In failure cases, the model often misinterprets the number or structure of blocks at the outset, causing errors to propagate throughout the reasoning process and ultimately affecting the final output.


\textbf{Performance degrades steeply from perception-level tasks to strategic multi-step planning, exposing VLMs’ limits in cross-modal reasoning and temporal integration.}
Experimental results reveal a typical hierarchical difficulty distribution across task levels, as shown in Figure \ref{fig:Radar6Models}, where Level 1 exhibits the lowest difficulty while Level 4 demonstrates the highest complexity. This progression highlights the limitations of current VLMs in tasks that transition from perceptual understanding to strategic planning. The performance degradation suggests that as tasks evolve from basic perception (e.g., object recognition) to advanced planning (e.g., multi-step reasoning), VLMs encounter challenges in effectively integrating multimodal information and executing systematic cognitive processes. Potential factors include insufficient contextual reasoning capacity, limited cross-modal alignment precision, and inadequate temporal dependency modeling in complex decision-making scenarios.
In our experiments, the tasks are organized into four tiers of increasing complexity (Level‑1 through Level‑4). Each successive level introduces additional blocks, more intricate spatial relationships, and higher-order reasoning requirements, thereby posing a progressively greater challenge for vision-language models.

\section{Conclusion}
\vspace{-12pt}
In this paper, we introduced PhyBlock, a novel benchmark for evaluating the scaling of cognitive skills in robotic 3D block assembly tasks. PhyBlock features a four-level difficulty scale, ranging from basic perception to advanced spatial planning, which allows for progressively challenging models in their cognitive abilities. The benchmark evaluates models based on three critical dimensions: partial task completion, failure diagnosis, and planning robustness. We applied PhyBlock to evaluate a range of state-of-the-art VLMs, providing a detailed analysis of their performance across these dimensions. Our results highlight both the strengths and weaknesses of these models, offering insights into their capabilities and areas for further improvement in handling complex, multi-step tasks.

\section{Acknowledgments}
This work is supported by National Key Research and Development Program of China (2024YFE0203100), National Natural Science Foundation of China (NSFC) under Grants No.62476293, National Postdoctoral Program for Innovative Talents under Grant Number BX20250379, China Postdoctoral Science Foundation under Grant Number 2025M771521, and General Embodied AI Center of Sun Yat-sen University.


\bibliographystyle{abbrv}
\bibliography{main}


\clearpage
\appendix

\section{Detailed Dataset Collection Process}
\label{app:data_cons}

\subsection{Construction of Modular 3D Block Simulation Assets}
\label{app:generate_assets}
\input{tables/block_survey}
Current research in embodied intelligence faces significant limitations in constructing simulation environments for block assembly tasks. Although several prior works have investigated manipulation of block-like objects, most of them do not release the core simulation assets. Existing open-source implementations are constrained by fixed shapes and predefined connection mechanisms, limiting their applicability in studies requiring generalization.
To systematically investigate embodied agents' capabilities in long-horizon planning and spatial reasoning within diverse block assembly scenarios, we develop a modular asset library of 3D blocks with geometrically extensible structures. This asset library is built using rigid-body modeling under physical constraints, supporting both compositional geometry and interactive simulation.

To ensure the constructed simulation environment aligns with real-world design patterns, we conducted a comprehensive survey of popular commercial block kits. As summarized in Table \ref{tab:block_survey}, we analyzed core attributes—such as shape categories, color variations, pattern types, and connection mechanisms—across different mainstream brands. These factors play a critical role in shaping the perception, manipulation, and strategy learning of embodied agents in complex block-building tasks.

Based on the above preliminary analysis, we adopt a hierarchical design principle in constructing the block asset library: \textbf{1) Geometric Primitives}: The library includes eight types of ISO-standard geometric shapes, such as cubes, cuboids, and triangular prisms, which together cover over 90\% of the basic forms found in commercial kits. \textbf{2) Color System}: We adopt five highly distinguishable colors—red, yellow, blue, green, and orange—to ensure visual clarity and perceptual diversity.

\begin{figure*}[t]
  \centering
   \includegraphics[width=1.0\linewidth]{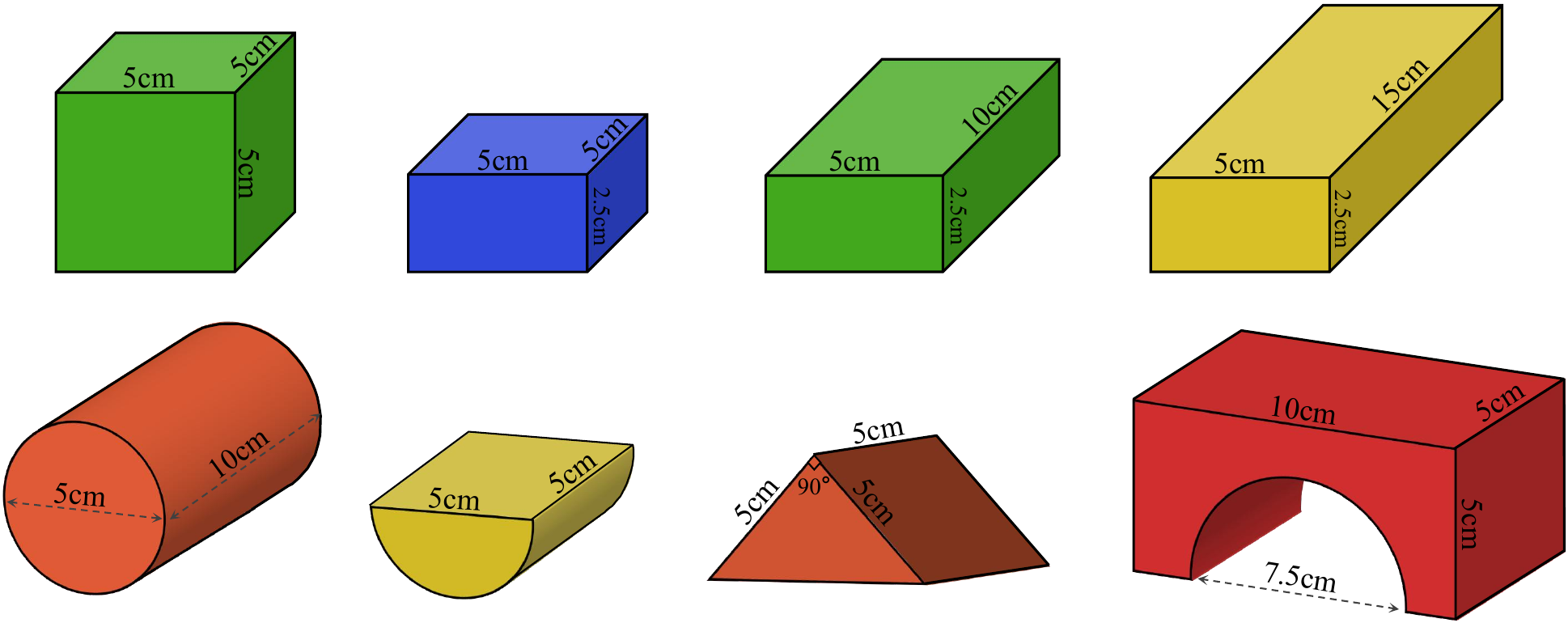}
   \caption{Basic Styles and Dimensional Specifications of 3D Simulated Block Models}
   \label{fig:blocks}
\end{figure*}

We use Blender as the primary platform for building our simulation assets, consisting of two stages: 
\textbf{1) Geometric Modeling}: Parameterized models are constructed in Blender using a 5 cm as the base unit. Boolean operations are applied to generate the eight standard geometric shapes.
\textbf{2) Physical Material Modeling}: The models are imported into Blender 3.4, where rigid-body dynamics are configured. We simulate ABS plastic properties by setting a friction coefficient of $\mu = 0.35$ and a density of $\rho = 1.04\ \mathrm{g/cm^3}$. To enhance surface detail and edge fidelity, normal mapping and edge subdivision techniques are applied. The final block asset style and size are shown in the figure \ref{fig:blocks}.

\subsection{Construction Pipeline of Block Assembly Scenes}
\label{app:generate_scenes}

Based on the aforementioned simulated block kit, we construct a variety of 3D block-building scenarios with different styles and levels of difficulty by composing individual blocks into diverse configurations. This process involves four key steps:

\textbf{(1) Collecting Example Images of Scene Styles.}
We curated a diverse set of block assembly images from the internet as inspiration and references for designing our simulation scenarios.

\textbf{(2) Manual Construction of Block Scenes.}
Each scene consists of multiple blocks of different types. To ensure high-quality dataset generation, it is essential to precisely annotate the spatial position, orientation, and topological dependencies among the blocks. We adopt a manual annotation pipeline to label the pose and relational structure of each block in a scene. All annotations are stored in structured \texttt{JSON} format. An example of such an annotation is illustrated in Figure~\ref{fig:JSON}.

\begin{figure}[t]
  \centering
   \includegraphics[width=1.0\linewidth]{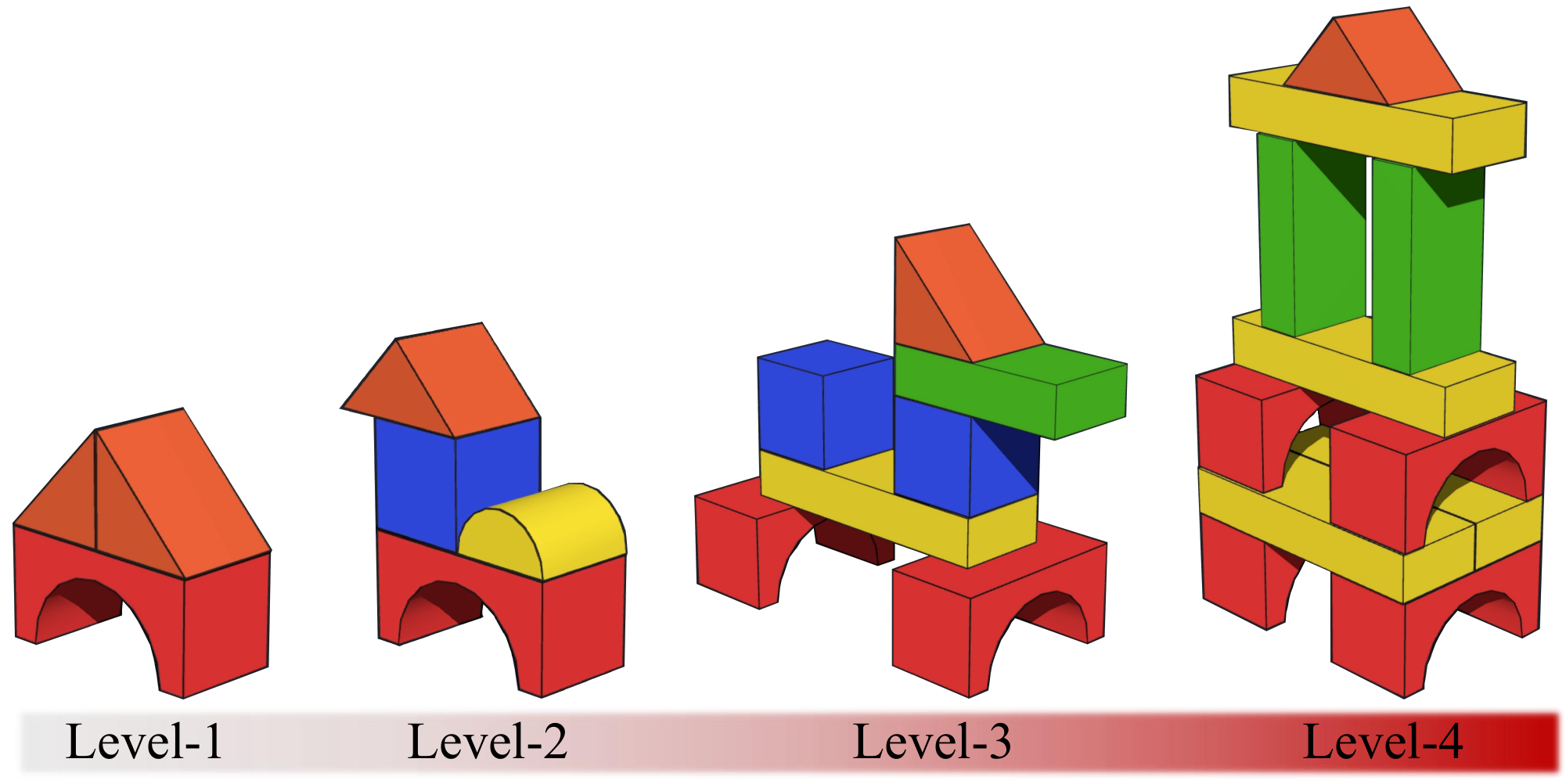}
   \caption{Illustration of Difficulty Levels in Block Assembly Tasks}
   \label{fig:levels}
\end{figure}

\begin{figure}[htbp]
  \centering
  \includegraphics[width=\linewidth]{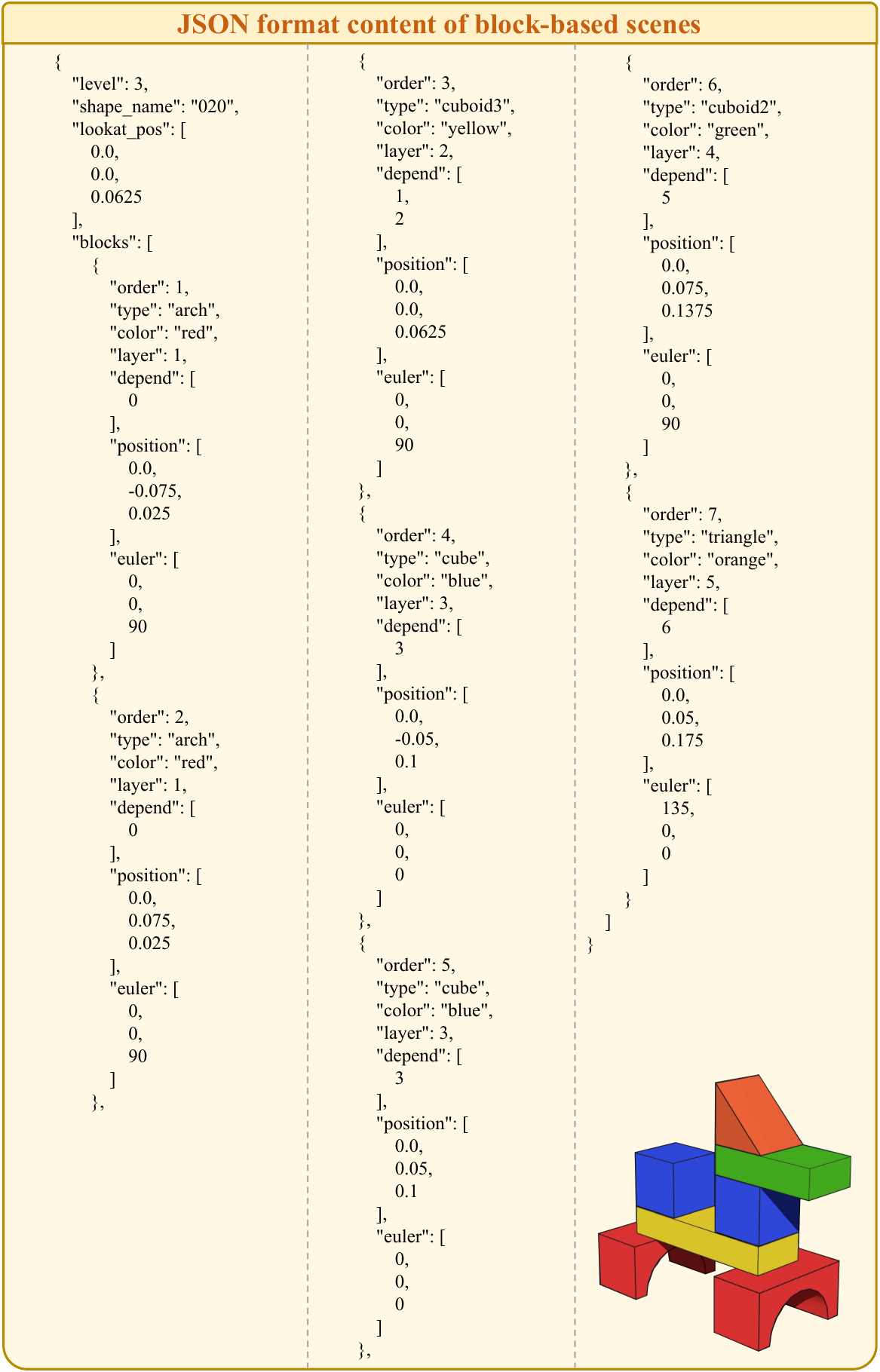}
  \caption{JSON File Content for a Sample Block Assembly Scene}
  \label{fig:JSON}
\end{figure}

\textbf{(3) Data Augmentation and Difficulty-Level Classification.}
We first manually constructed 150 unique block scenes and then applied geometric augmentations such as rotation to expand the dataset to 400 scenes. The final dataset is categorized into four difficulty levels: Level-1, Level-2, Level-3, and Level-4, containing 36, 121, 138, and 119 scenes respectively, as illustrated in Figure~\ref{fig:levels}. Note that Level-1 and Level-2 exhibit partial overlap.

\textbf{(4) Simulation.}
Using the \texttt{Genesis} platform, we rendered the block-building scenes under 6 varying background environments and camera viewpoints. The resulting multi-view images serve as the basis for subsequent question-answering data design.

Moreover, the difficulty of each block assembly scene is significantly influenced by four key factors: the number of blocks involved, the diversity of block types, the variety of colors, and the depth of the final assembly hierarchy. As illustrated in Fig.\ref{fig:maxavg}, we provide a quantitative analysis of these four dimensions across different difficulty levels. Specifically, Fig.\ref{fig:maxavg}(a) presents the maximum values observed in each dimension, while Fig.~\ref{fig:maxavg}(b) reports the corresponding mean values. By comparing the two radar charts, we observe a clear upward trend across all dimensions as the difficulty level increases. This progressive pattern demonstrates the effectiveness and rationality of our dataset design in stratifying task complexity. Such a difficulty-aware structure is essential for benchmarking model performance across varying levels of planning and reasoning challenges.
\input{Figs/Appendix_A/AssemPlan_Blocks_MaxAvg}

\subsection{Data Format and Structural Representation of Block Scenes}

Each constructed block scene is stored in a structured \texttt{JSON} format, which encapsulates both the high-level scene attributes and the fine-grained block-wise specifications. This structured format ensures that the dataset can be easily parsed and utilized in simulation platforms or learning algorithms. An example is shown below:

\begin{itemize}
    \item \textbf{"level"}: An integer indicating the difficulty level of the block scene, ranging from 1 (easiest) to 4 (most complex).
    \item \textbf{"shape\_name"}: A unique identifier string assigned to each scene configuration.
    \item \textbf{"blocks"}: A list of dictionaries, each representing an individual building block in the scene. The detailed fields are:
    \begin{itemize}
        \item \textbf{"order"}: A unique integer index indicating the ID of the block within the scene.
        \item \textbf{"type"}: The geometric category of the block (e.g., \texttt{"cube"}, \texttt{"cuboid2"}, \texttt{"arch"}, \texttt{"triangle"}), consistent with the primitives defined in our asset library.
        \item \textbf{"color"}: A categorical string indicating the block's color.
        \item \textbf{"layer"}: An integer representing the vertical level or stacking layer of the block, where a higher value implies a physically higher placement in the structure.
        \item \textbf{"depend"}: A list of integer indices referencing other blocks that this block is dependent on (i.e., those that must be placed before this block in the stacking process). These dependencies form a directed acyclic graph (DAG) that defines the scene's topological constraints.
        \item \textbf{"position"}: A 3D vector \texttt{[x, y, z]} specifying the center position of the block in the world coordinate frame, expressed in meters.
        \item \textbf{"orientation"}: A 3D vector \texttt{[roll, pitch, yaw]} defining the block's orientation using Euler angles in degrees, following the XYZ convention.
    \end{itemize}
\end{itemize}

This format provides a comprehensive and interpretable representation of the scene configuration, facilitating reproducibility, rendering, and task reasoning. Notably, the inclusion of topological dependencies (\texttt{"depend"}) allows for accurate reconstruction of the assembly process, which is crucial for downstream embodied manipulation and reasoning tasks. Figure~\ref{fig:JSON} illustrates an example of block scene annotation encoded in structured JSON format.

\subsection{Dataset Construction for Physical Understanding VQA}
\label{app:generate_VQA}
\subsubsection{Large Language Model-based VQA Dataset Generation}

\begin{figure}[t]
  \centering
   \includegraphics[width=1.0\linewidth]{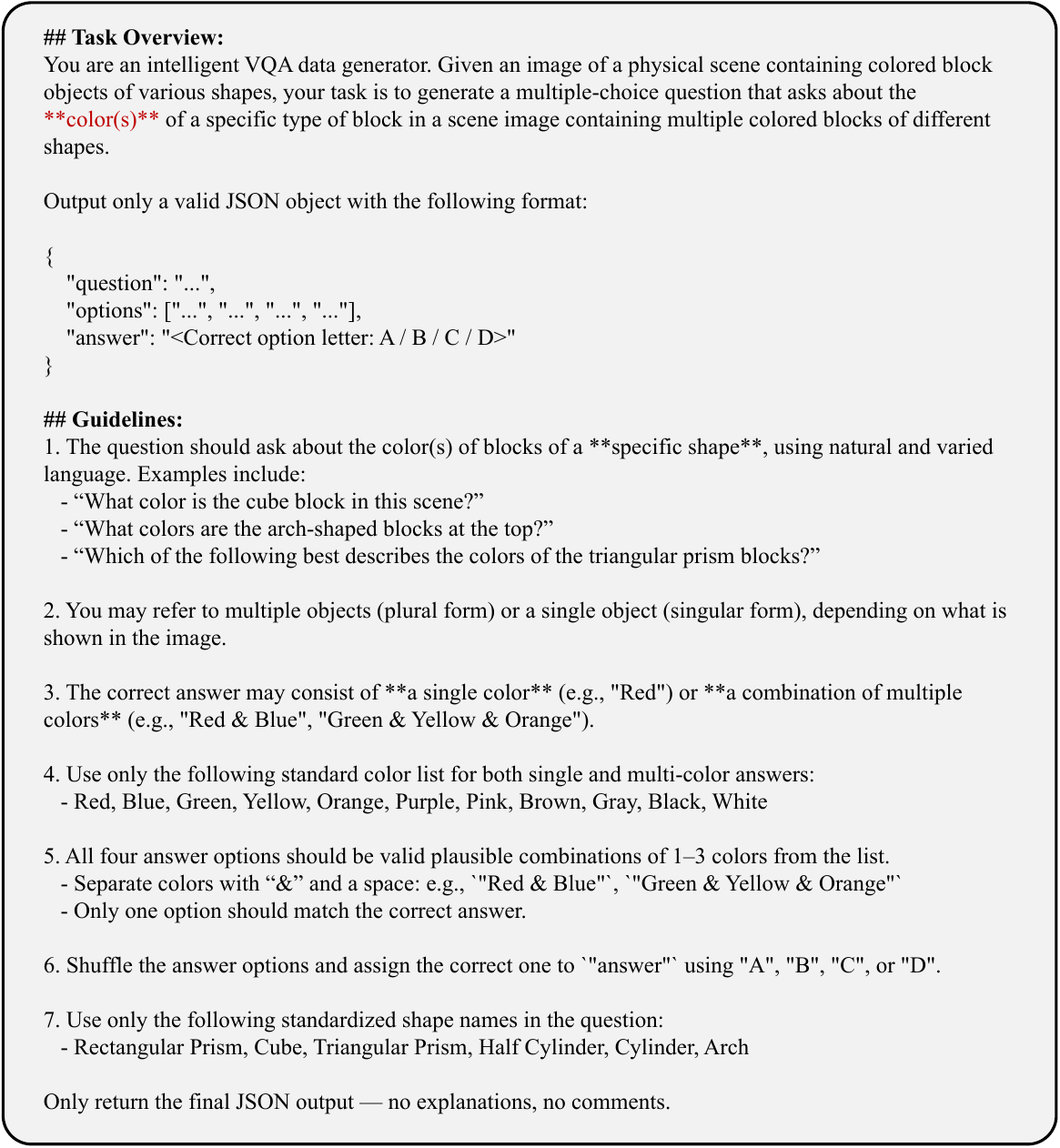}
   \caption{Example Prompt Design for the \textit{Color} Question Type}
   \label{fig:color}
   \vspace{-12pt}
\end{figure}

To complement our block assembly scenes with question-answering tasks, we leverage a large language model (LLM) to generate a suite of Visual Question Answering (VQA) samples targeting fundamental perception skills such as color recognition and object counting. This automated data generation process consists of three main stages:

\textbf{(1) Prompt Engineering.}
We first design a series of detailed prompts that instruct the LLM to generate high-quality, scene-aware VQA samples. Each prompt specifies the desired question type (e.g., \textit{Color}, \textit{Number}) and provides guidelines on phrasing, semantic clarity, and answer format. These prompts also include visual context descriptions to ensure the questions are relevant and grounded in the associated block scene images. Figure~\ref{fig:color} illustrates a prompt example for a \textit{Color}-type question.

\textbf{(2) LLM-Based Data Generation.}  
Using the curated prompts, we interact with a commercial LLM API---specifically, the latest GPT-4o model---to produce candidate VQA entries for each block scene. Each entry includes a scene ID, a natural language question, four multiple-choice answer options, and the correct answer label.

\textbf{(3) Human Validation and Quality Assurance.}  
To ensure the reliability and fairness of the generated dataset, we conduct a rigorous manual verification process. Three human annotators independently review each generated question three times, correcting semantic inaccuracies, verifying visual consistency, and rebalancing answer distributions. This intensive validation effort spans over 1{,}000 hours, resulting in a high-quality VQA dataset with accurate answers, diverse question formulations, and well-balanced choice distributions across scenes.

\begin{figure}[t]
    \centering
    \begin{subfigure}{1.0\textwidth}
        \centering
        \includegraphics[width=\linewidth]{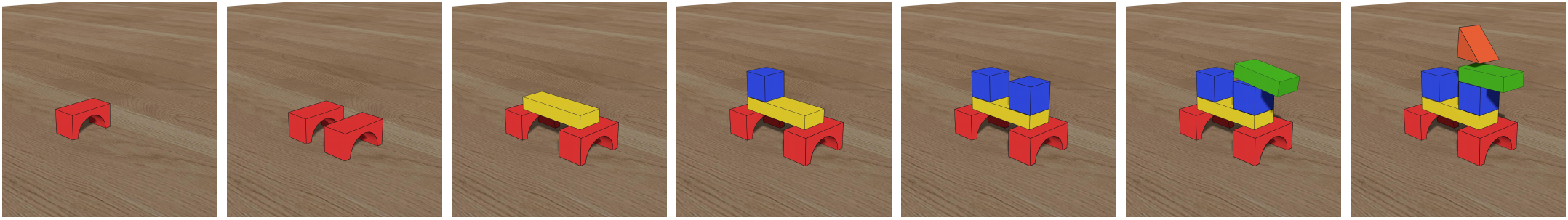}
        \caption{Step-by-Step Construction Rendered by the Simulation Engine}
    \end{subfigure}
    \vskip 1em  
    \begin{subfigure}{1.0\textwidth}
        \centering
        \includegraphics[width=\linewidth]{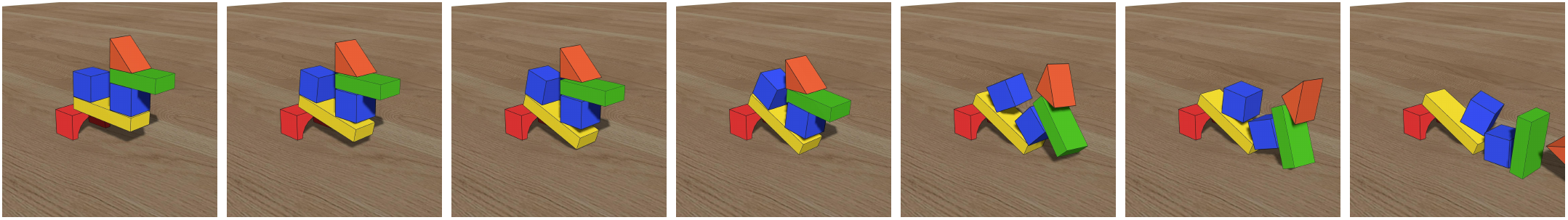}
        \caption{Dynamic Perturbations Rendered by the Simulation Engine}
    \end{subfigure}
    \caption{Visual Data Generated by Simulation Engine-based Methods}
    \label{fig:EngineBased}
\end{figure}

\subsubsection{Simulation Engine-based VQA Data Generation}
While large language models (LLMs) are well-suited for generating simple, static visual question answering (VQA) samples, more complex question types—such as \textit{Predictive} and \textit{Counterfactual} reasoning—require dynamic scene understanding grounded in physical interactions. To this end, we develop a simulation-driven VQA data engine based on the \texttt{Genesis} platform to support physically grounded question generation. The process involves three main components:

\textbf{(1) Question Template Design.}  
We first construct diverse and flexible question templates with the assistance of GPT-4o, enabling coverage across various reasoning scenarios. For instance, for the \textit{Predictive} question category, we design multiple paraphrased templates to elicit responses about scene evolution following a specific perturbation. Sample templates include:
\begin{itemize}
    \item What is likely to happen if the \{color\} \{type\} block is taken away?
    \item How will the scene change if the \{color\} \{type\} block is removed?
    \item What consequences might follow the removal of the \{color\} \{type\} block?
    \item Suppose the \{color\} \{type\} is taken out—what happens then?
\end{itemize}
Here, \texttt{color} spans \textit{Red, Blue, Green, Yellow, Orange}, and \texttt{type} spans \textit{Rectangular Prism, Cube, Triangular Prism, Half Cylinder, Cylinder, Arch}.

\textbf{(2) Simulation-based Scene Perturbation.}  
Based on existing JSON scene annotations, we randomly apply physically plausible perturbations—such as removing a block—to instantiate specific question templates. Each perturbed scene is then reconstructed and simulated within the Genesis engine. The resulting scene evolution is rendered as a short video (in \texttt{.mp4} format), capturing the dynamic changes. To support VQA input-output formatting, selected video frames are exported as \texttt{.png} images and used either as question prompts or as visual multiple-choice options.

\textbf{(3) Human Verification.}  
To ensure high data quality, three annotators performed three rounds of thorough manual verification, spending over 500 cumulative hours. This process ensured correctness of the VQA samples, balanced question-type and option-type distributions, and eliminated potential annotation noise or ambiguities.

This simulation-driven pipeline enables us to systematically generate physically grounded VQA samples, extending beyond the static-image domain to support robust understanding of cause-effect dynamics in structured environments.

\subsection{Data Format and Structural Design for VQA Tasks}
\begin{figure}[t]
  \centering
   \includegraphics[width=1.0\linewidth]{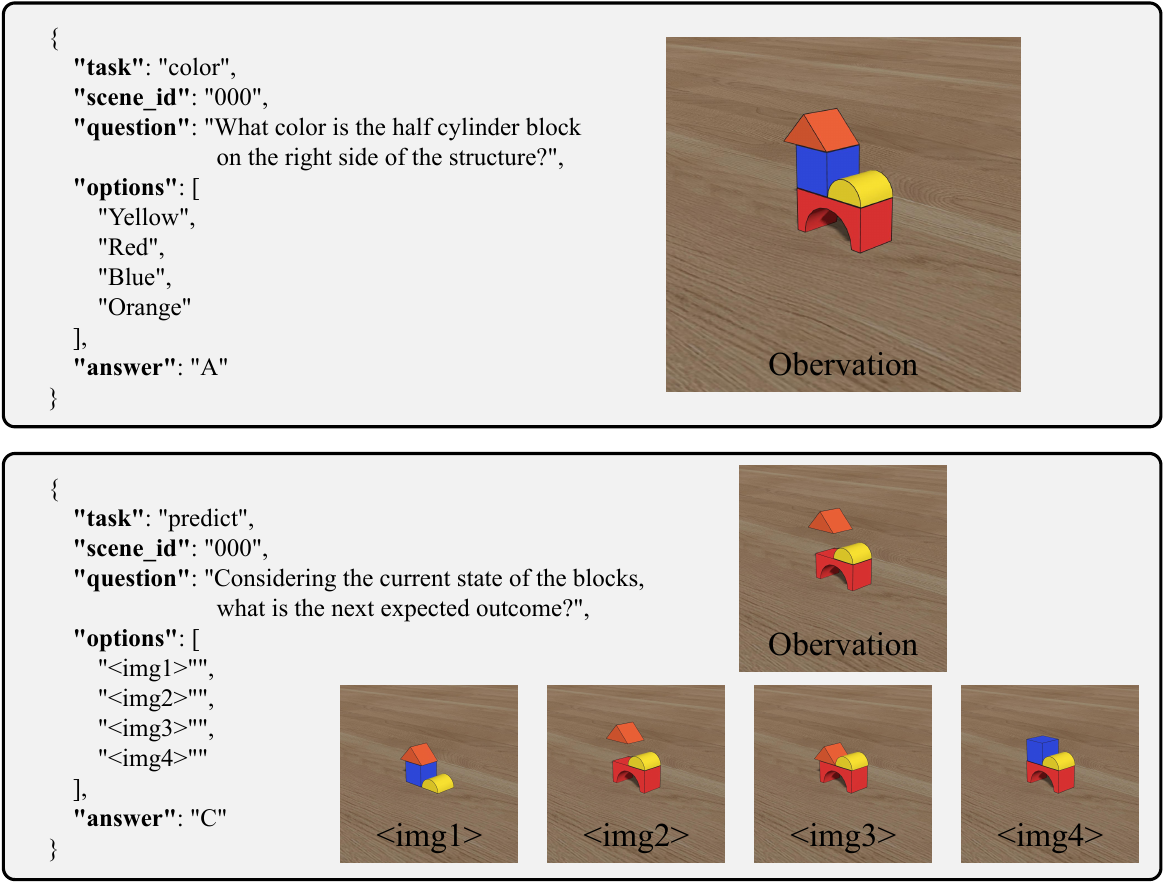}
   \caption{VQA Format Examples for \textit{Color} and \textit{Predict} Subtasks}
   \label{fig:JSON2}
\end{figure}

To support various types of Visual Question Answering (VQA) tasks in our block assembly benchmark, we design a unified and lightweight data format that facilitates model training, evaluation, and interpretability. Each VQA sample is stored in a structured JSON format, as shown in Figure \ref{fig:JSON2}, containing the following key elements:

\begin{itemize}
  \item \textbf{scene\_id}: A unique identifier for the associated block scene.
  \item \textbf{question}: A natural language question grounded in the visual content of the scene.
  \item \textbf{options}: A list of four candidate answers, typically labeled A–D. The modality of the options varies across different subtasks. For standard tasks such as \textit{Color}, \textit{Number}, or \textit{Shape}, the options are textual descriptions. However, for more complex reasoning subtasks—including \textit{dynamic}, \textit{counterfactual}, and \textit{ordering}—the answer choices are rendered as images corresponding to possible scene evolutions or structural outcomes. In such cases, the options are denoted as \texttt{<img1>}, \texttt{<img2>}, \texttt{<img3>}, and \texttt{<img4>}, referring to the paths of the candidate image files.

  \item \textbf{answer}: The correct answer's index or label, aligned with one of the provided options.
\end{itemize}

This design ensures extensibility across a wide range of question categories—such as object attributes, spatial relations, numerical counting, and dynamic reasoning—while maintaining consistency across annotations. This multi-modal design across 16 subtasks enables our dataset to capture a broad spectrum of physical and semantic reasoning challenges.

\clearpage
\section{More Details On The Setup}
\label{app:eval}

\subsection{Assembly Step Evaluation via AOV Network}
\label{sub:aov}

\textbf{Core Idea:}  
In this work, we design a model that performs multimodal reasoning over a combination of inputs: a target reference image, candidate block images, and a natural language instruction describing the assembly process. The model is expected to integrate both visual and linguistic information to identify the correct building components from the candidates and generate a complete, ordered sequence of assembly actions. 

To systematically assess the model's performance on this task, we propose an evaluation algorithm based on the \textit{Assembly Order Validation} (AOV) network. The goal of this algorithm is to verify the correctness of each predicted assembly step by comparing it with a ground-truth assembly sequence. Specifically, the AOV network determines whether each predicted operation adheres to the reference construction order, thus evaluating the model's ability to understand and respect the sequential dependencies inherent to the assembly process. By aligning the predicted sequence with the ground truth, the method enables the computation of standard evaluation metrics, including True Positives (TP), False Positives (FP), and False Negatives (FN), thereby quantifying the model’s reasoning accuracy, structural awareness, and robustness in task execution.

\textbf{Algorithmic Procedure:}  
To comprehensively evaluate a model’s reasoning ability in the block assembly task, we develop a sequence-matching algorithm that compares the predicted assembly sequence against the ground-truth order, as detailed in Algorithm~\ref{algo:block_eval}. The algorithm begins by initializing all ground-truth blocks with an “unplaced” status and resetting the match flags for all predicted blocks. It then iterates over each predicted block placement, searching for a matching target in the ground-truth sequence that: (1) has not yet been placed, (2) has exactly the same geometric properties (e.g., position and pose), and (3) satisfies topological feasibility in the assembly structure.

Once a valid match is found, the corresponding ground-truth block is marked as placed, and the predicted block’s match index is recorded for subsequent metric computation. Upon completion of the matching process, the algorithm computes the following key metrics:  
- \textbf{TP} (True Positives): the number of correctly matched predicted blocks,  
- \textbf{FP} (False Positives): the number of unmatched or incorrectly matched predicted blocks, and  
- \textbf{FN} (False Negatives): the number of ground-truth blocks not matched by any prediction.  

Based on the proposed evaluation algorithm, the computed TP, FP, and FN quantify the prediction performance for each individual block assembly scene. These values are further used to calculate the \textbf{precision}, \textbf{recall}, and \textbf{F1-score} at the scene level. While these metrics effectively capture local reasoning performance for individual samples or task-level instances, we also adopt the \textbf{micro-averaged F1-score} (Micro-F1) to aggregate performance across all samples, thereby providing a comprehensive evaluation of the model's global assembly reasoning capability.
\input{tables/AOV_algo}

\subsection{Two Evaluation Settings for 3D Block Assembly Step Planning}
\label{sub:B2}
\begin{figure}[b]
  \centering
   \includegraphics[width=1.0\linewidth]{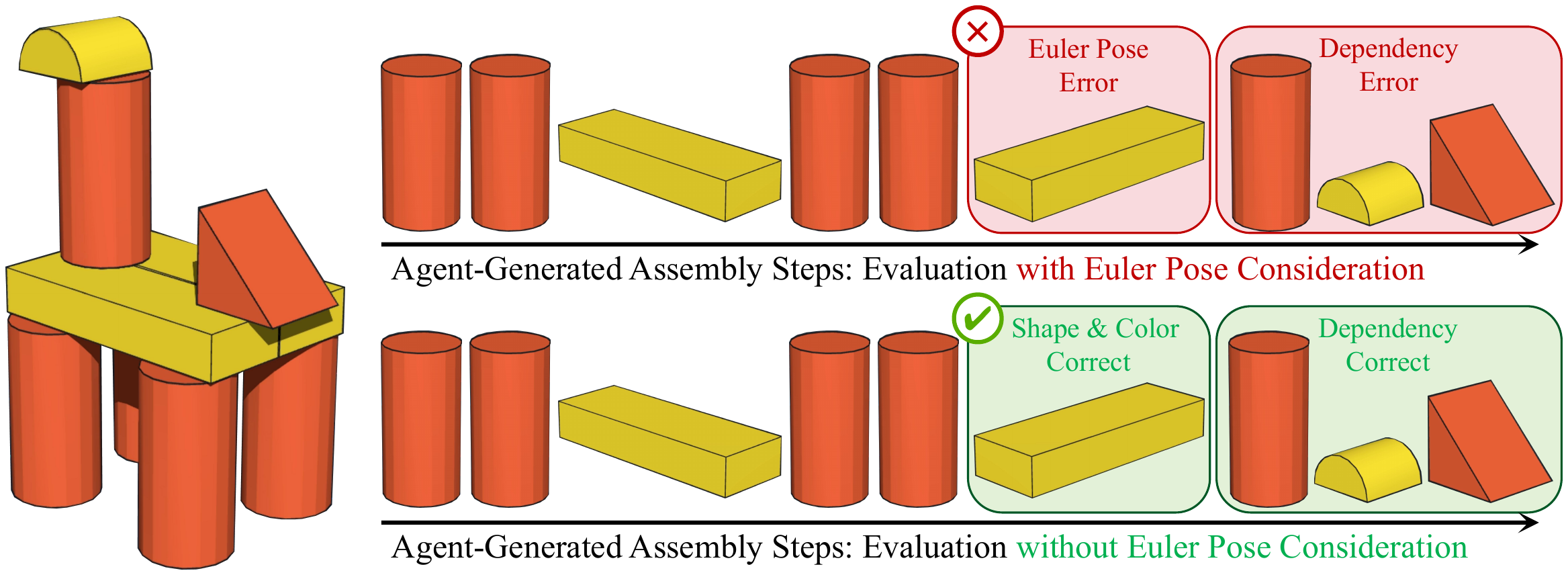}
   \vspace{-4mm}
   \caption{The situation under two evaluation settings \textcolor{red}{A} and \textcolor{blue}{B}}
   \label{fig:Scene_Giraffe_Setting}
   \vspace{-4mm}
\end{figure}

We propose a two-dimensional evaluation framework to systematically assess a model’s scene understanding and 3D block assembly capabilities. As illustrated in Fig.~\ref{fig:inference_setting}, the model is required to solve two hierarchical tasks: (1) select the necessary blocks (in terms of shape, color, and pose) from a candidate pool based on a reference image, and (2) generate a step-by-step assembly plan. To rigorously quantify model performance, we introduce two complementary evaluation paradigms:

\paragraph{\textcolor{red}{A.} Pose-Constrained Evaluation (with orientation Consideration).}
Under this strict setting, the model must produce an assembly sequence that exactly matches the ground truth orientation,  as detailed in Algorithm \ref{algo:block_eval}. Each predicted step is considered correct only if the selected block’s shape, color, and orientation all match the corresponding ground-truth attributes. This paradigm emphasizes the model’s geometric reasoning ability, particularly its precision in understanding 3D rotational configurations. As shown in Fig.~\ref{fig:Scene_Giraffe_Setting}, even a minor pose error in the yellow cuboid leads to structural failure in subsequent steps, highlighting the setting’s sensitivity to long-range planning inconsistencies. This strict constraint enables a fine-grained evaluation of the model’s robustness in complex spatial reasoning tasks.

\paragraph{\textcolor{blue}{B.} Topology-Oriented Evaluation (without orientation Consideration).}
To decouple pose sensitivity from structural planning performance, this relaxed setting ignores the orientation differences. A predicted step is deemed correct if the shape and color of the selected block match those of the ground truth, regardless of its pose. This paradigm focuses on the model’s structural planning and task decomposition capabilities, effectively mitigating the impact of local pose inaccuracies on global evaluation. As shown in Fig.~\ref{fig:Scene_Giraffe_Setting}, although the yellow cuboid has an incorrect pose, it still functions as a supporting structure, allowing the assembly to proceed correctly. This setting is particularly well-suited for evaluating the model’s high-level reasoning and planning competence.

Together, these two evaluation protocols form a layered diagnostic framework that isolates and analyzes different dimensions of model ability. The pose-constrained evaluation probes geometric accuracy, while the topology-oriented evaluation targets structural reasoning. This hierarchical design provides a more nuanced and interpretable assessment of model performance, facilitating the identification of potential bottlenecks in assembly planning algorithms.

\subsection{Four Error Type Classifications for 3D Block Assembly Step Analysis}
\label{sub:errortype}

To better understand the limitations of multimodal large models in multi-step reasoning tasks for 3D block assembly, we introduce a structured error analysis framework that categorizes step-level mistakes into four types: \textbf{Incorrect orientation chosen}, \textbf{Dependency Not Met}, \textbf{Shape Not in GT}, and \textbf{Overflow}.

\paragraph{Incorrect Orientation Chosen.}  
This error occurs when the model correctly identifies the block type but predicts an incorrect orientation. Such rotation errors lead to misaligned placements despite the correct choice of block.

\paragraph{Dependency Not Met.}  
This type of error arises when the placement of a block fails due to missing or incorrectly positioned prerequisite blocks from earlier steps. As a result, spatial or structural dependencies required for proper placement are violated.

\paragraph{Shape Not in GT.}  
This error indicates that the model selects a block that does not belong to the ground-truth target set. It reflects a misidentification in shape or type, diverging from the intended assembly goal.

\paragraph{Overflow.}  
Overflow errors represent redundant actions, where the model attempts to place a block that has already been correctly positioned. These unnecessary placements lead to structural overbuild or duplication.

\begin{figure}[t]
  \centering
   \includegraphics[width=1.0\linewidth]{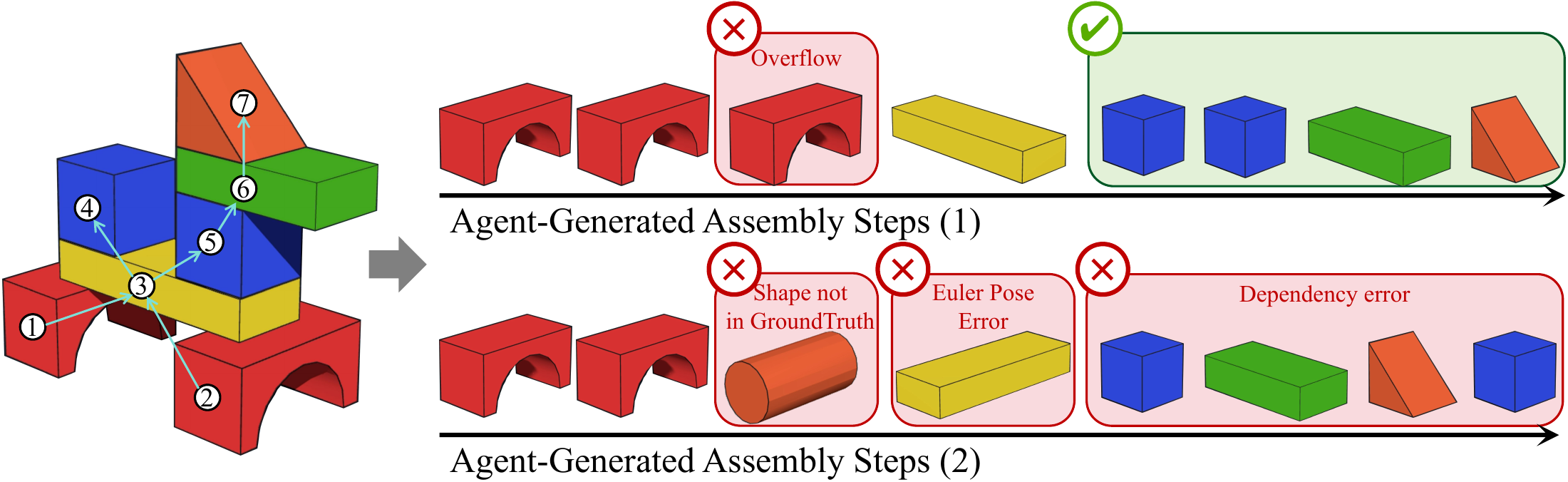}
   \vspace{-4mm}
   \caption{Four types of errors arising from the reasoning process}
   \label{fig:ErrorTypes}
   \vspace{-4mm}
\end{figure}

Figure~\ref{fig:ErrorTypes} presents illustrative examples for each of the defined error types.
Specifically, in Assembly Trace (1), Step 3 involves a redundant placement of the red arch block, which is categorized as an \textbf{Overflow Error}.
In Assembly Trace (2), Step 3 incorrectly selects an unnecessary orange cylinder that does not belong to the target block set, constituting a \textbf{Shape Not in GT} error.
In Step 4, although the yellow cuboid is the correct block, its predicted orientation does not match the ground-truth pose, thus falling under the \textbf{Incorrect Orientation Chosen} category.
Moreover, the failure to properly place the yellow cuboid in the second layer results in a cascade of \textbf{Dependency Not Meet} errors, as blocks in the third, fourth, and fifth layers rely on it for support and cannot be placed successfully.

This fine-grained evaluation paradigm not only pinpoints the root causes of performance discrepancies across models, but also provides a more interpretable perspective for understanding the underlying mechanisms of reasoning failures.

\clearpage
\subsection{Prompt Design for 3D Block Assembly Step Planning}

\begin{figure}[htbp]
  \centering
   \includegraphics[width=1.0\linewidth]{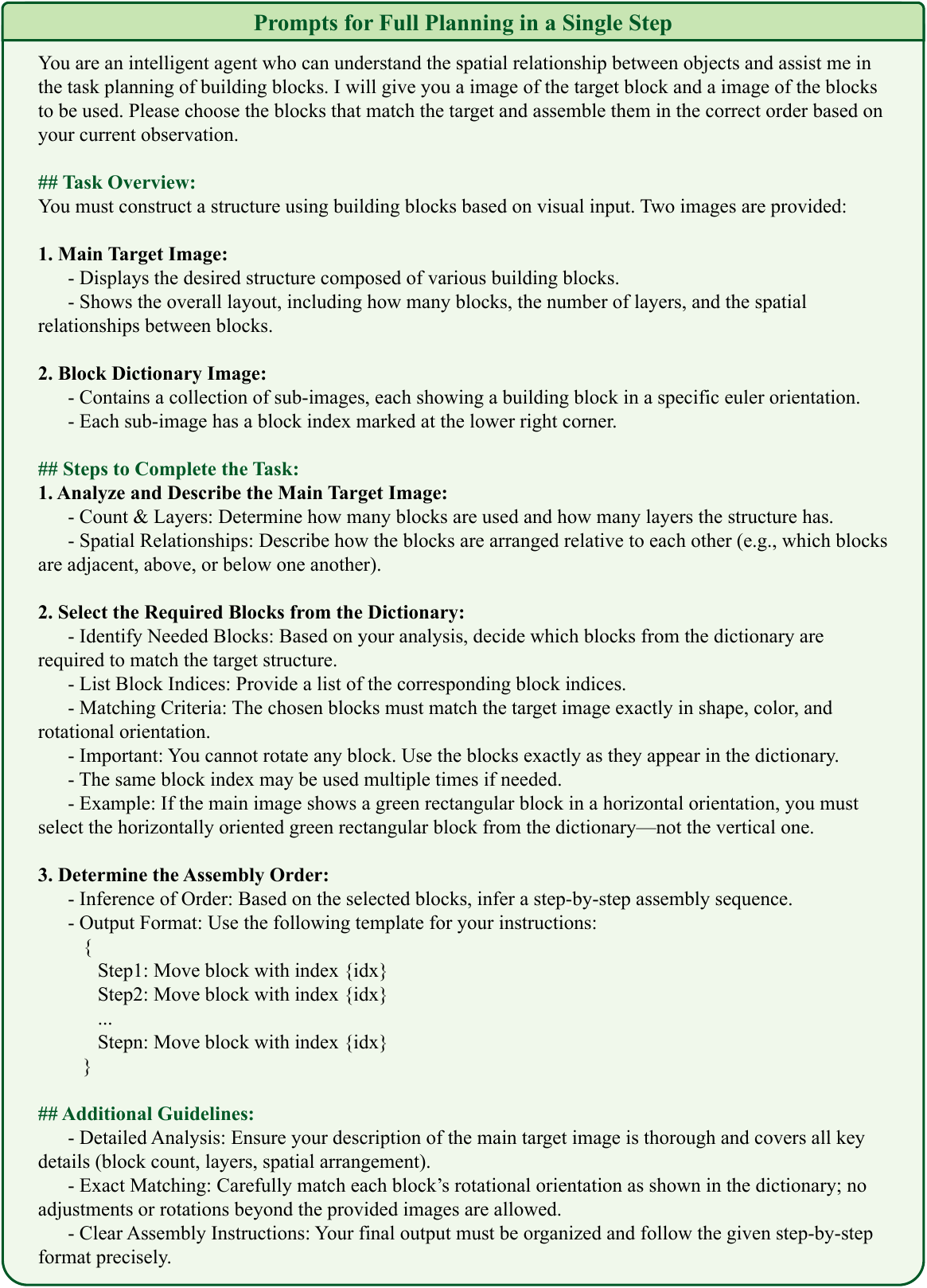}
   \caption{Prompts for Full Planning in a Single Step}
   \label{fig:prompt_full}
   \vspace{-12pt}
\end{figure}

\begin{figure}[htbp]
  \centering
   \includegraphics[width=1.0\linewidth]{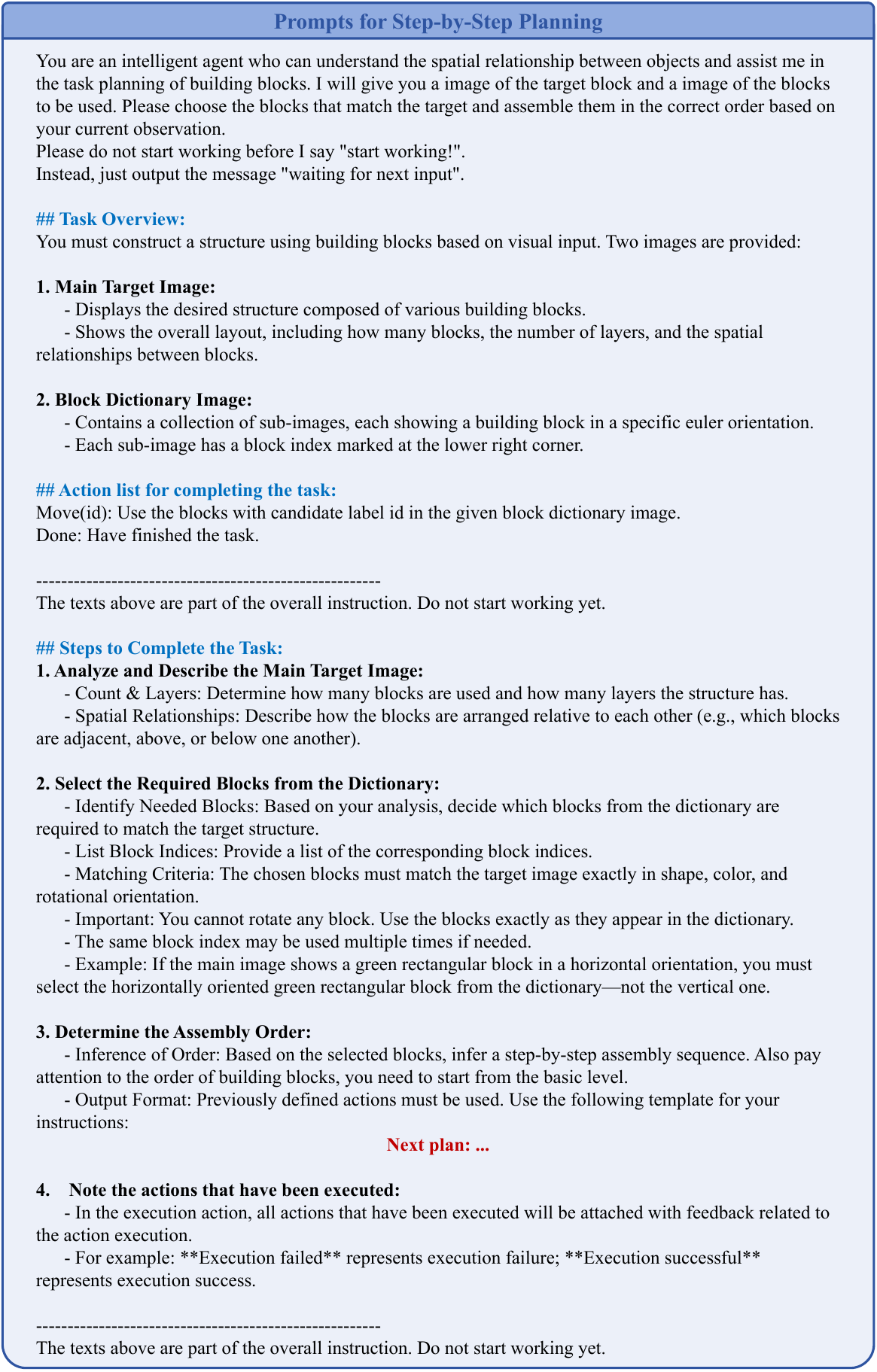}
   \caption{Prompts for Step-by-Step Planning}
   \label{fig:prompt_sbs1}
   \vspace{-12pt}
\end{figure}

\begin{figure}[htbp]
  \centering
   \includegraphics[width=1.0\linewidth]{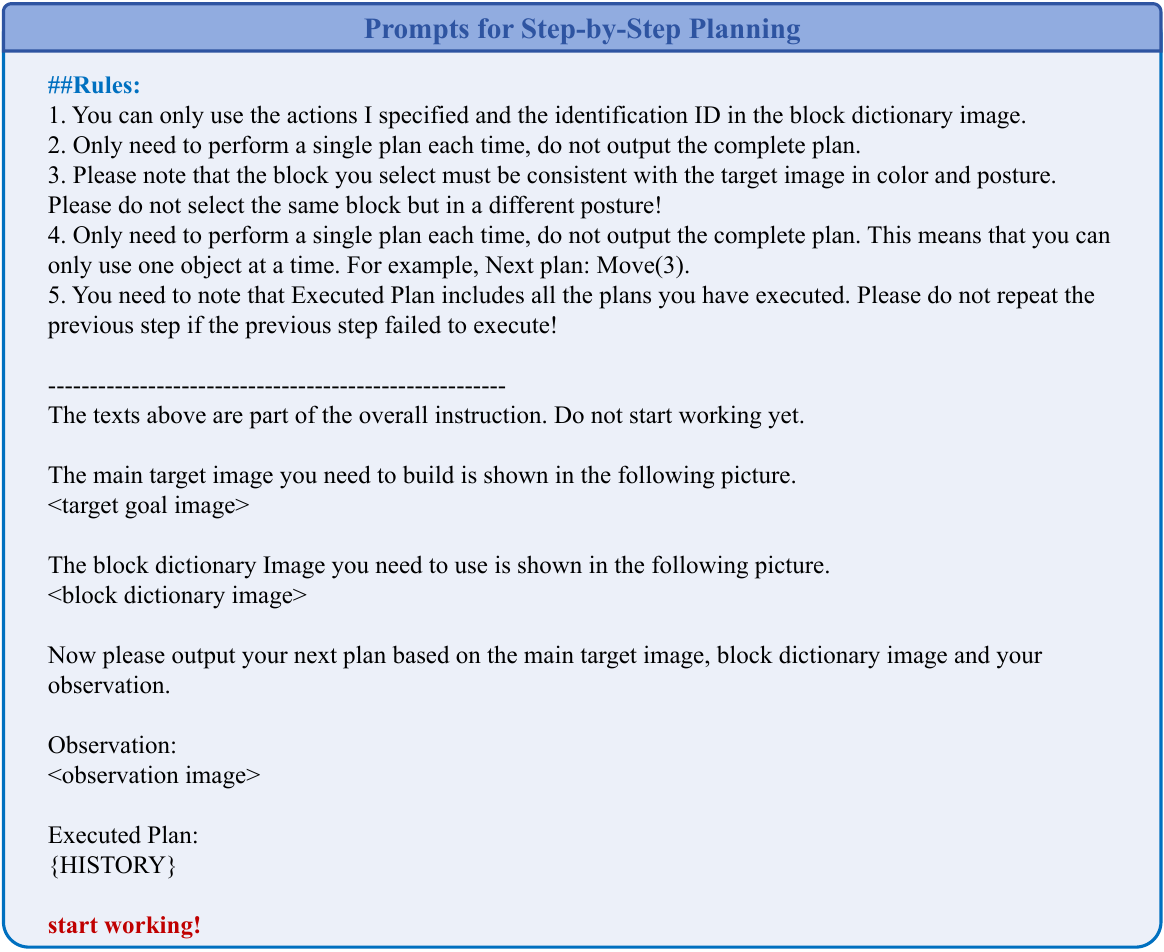}
   \caption{Prompts for Step-by-Step Planning}
   \label{fig:prompt_sbs2}
   \vspace{-4mm}
\end{figure}

\clearpage
\section{More Examples}
\label{app:example}

\subsection{Examples under Six Background Conditions}
\begin{figure}[htbp]
    \centering
    \includegraphics[width=1.0\linewidth]{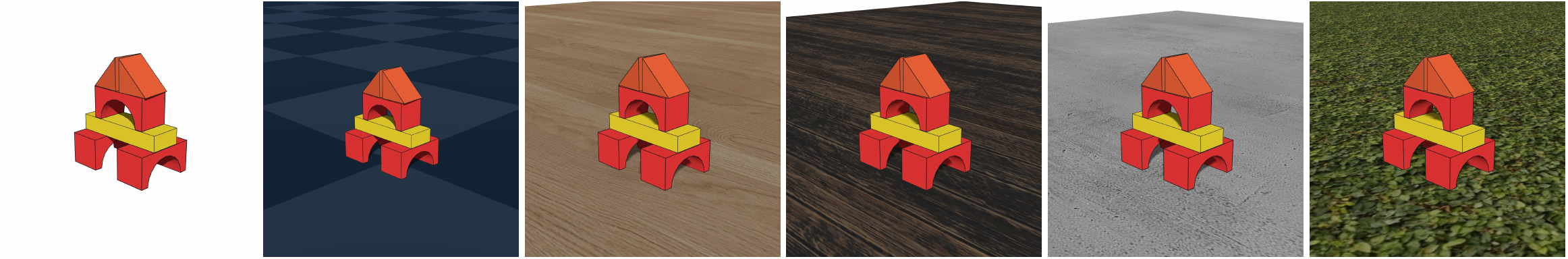}
    \caption{Block Assembly Scenes Across Six Environmental Backgrounds}
    \label{fig:6BackGround}
    \vspace{-12pt}
\end{figure}

\subsection{Examples from 3D Block Assembly Scenes}
\vspace{-12pt}
\begin{figure}[htbp]
    \centering
    \includegraphics[width=1.0\linewidth]{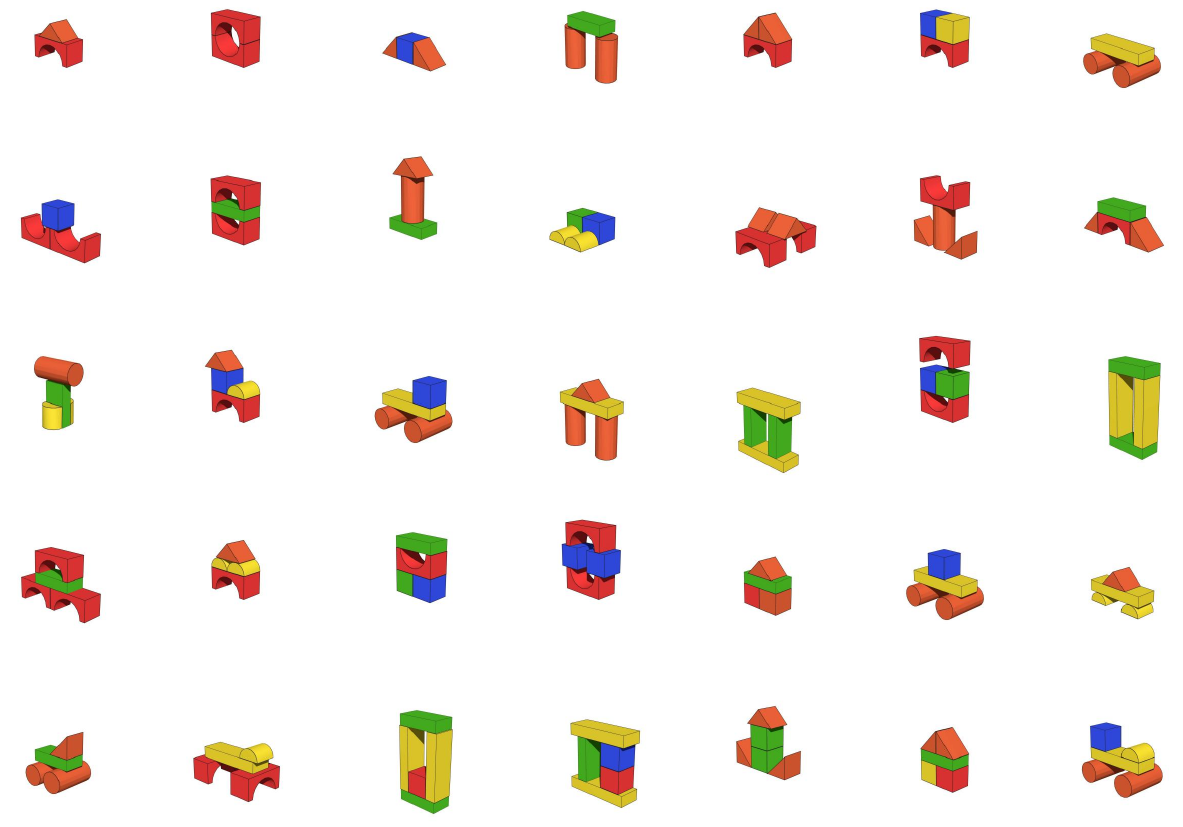}
    \caption{Partial Block Assembly Scenes at \textbf{Level-1} Difficulty}
    \label{fig:level1_clean_35}
    \vspace{-12pt}
\end{figure}

\begin{figure}[htbp]
    \centering
    \includegraphics[width=1.0\linewidth]{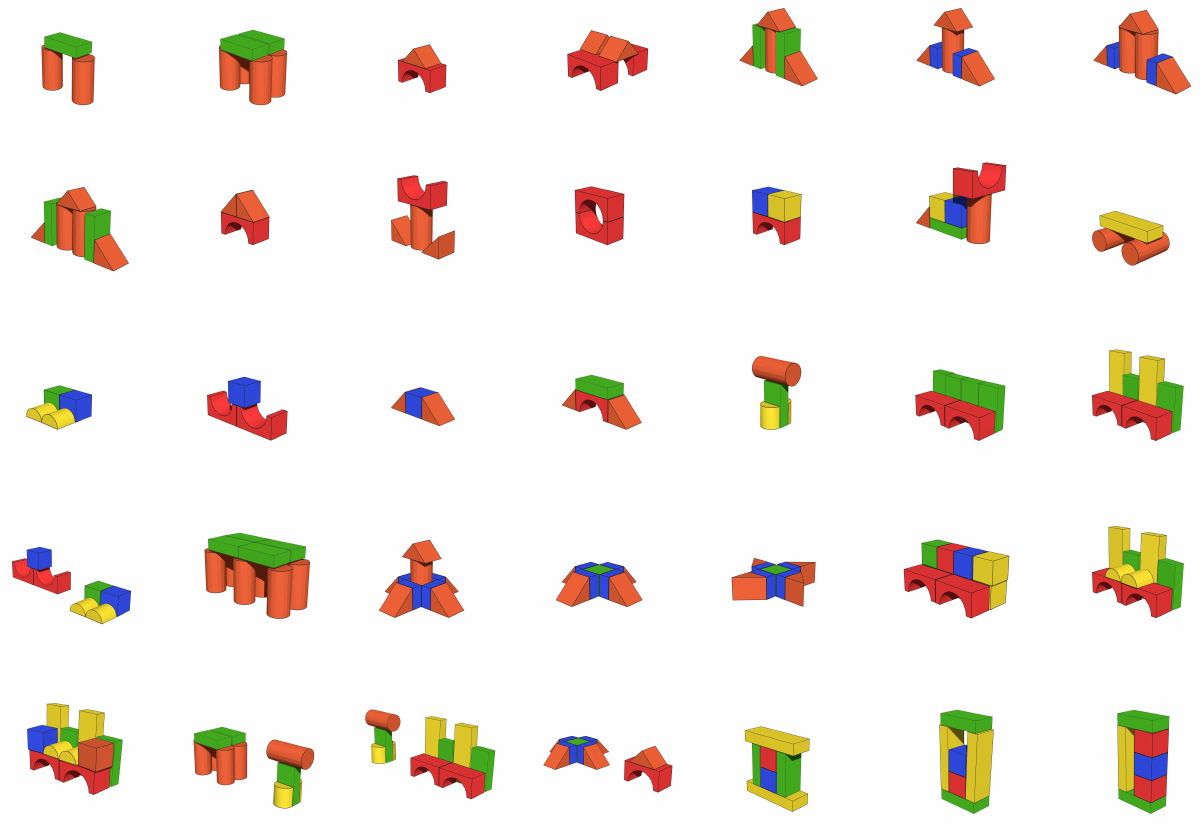}
    \caption{Partial Block Assembly Scenes at \textbf{Level-2} Difficulty}
    \label{fig:level2_clean_35}
    \vspace{-12pt}
\end{figure}

\begin{figure}[htbp]
    \centering
    \includegraphics[width=1.0\linewidth]{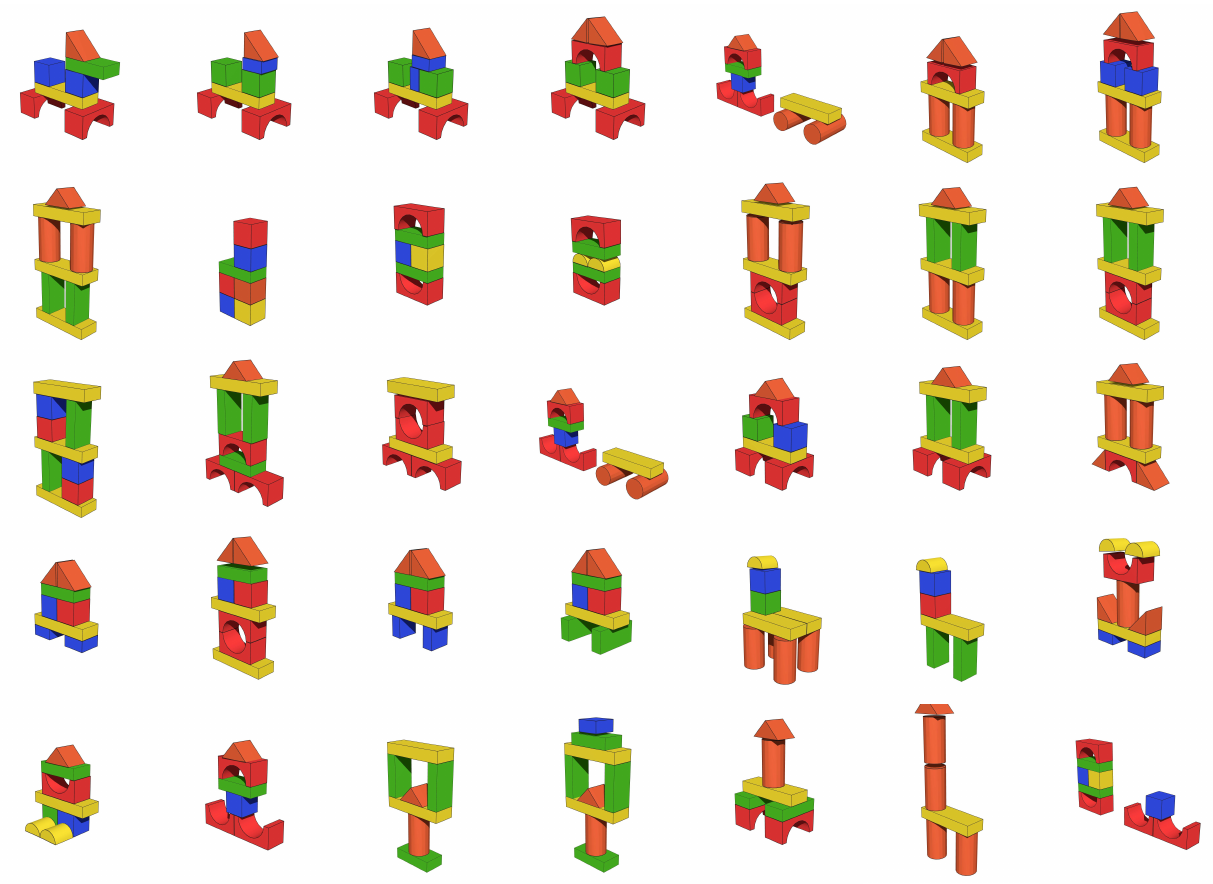}
    \caption{Partial Block Assembly Scenes at \textbf{Level-3} Difficulty}
    \label{fig:level3_clean_35}
    \vspace{-12pt}
\end{figure}

\begin{figure}[htbp]
    \centering
    \includegraphics[width=1.0\linewidth]{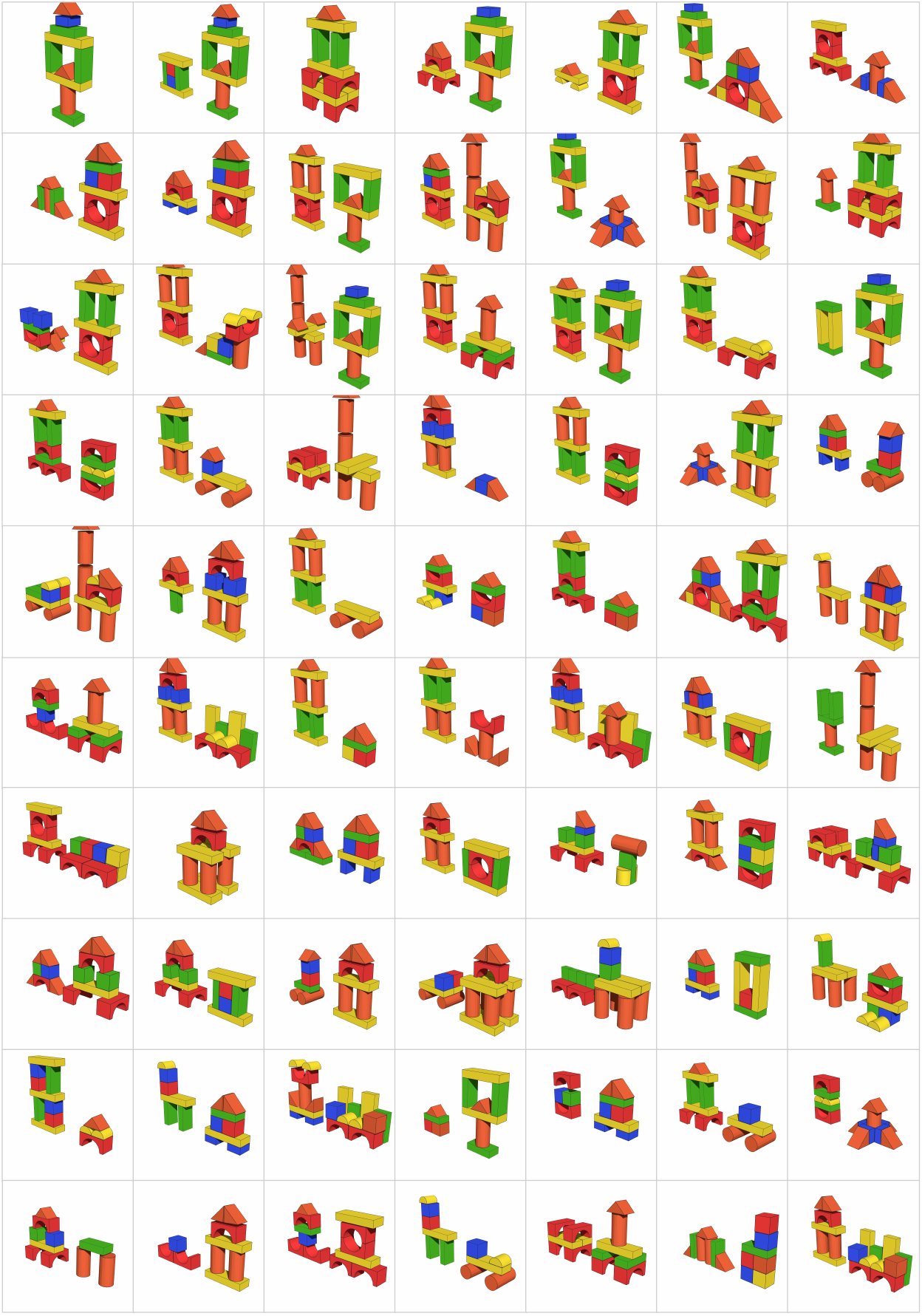}
    \caption{Partial Block Assembly Scenes at \textbf{Level-4} Difficulty}
    \label{fig:level4_clean_70}
    \vspace{-12pt}
\end{figure}

\clearpage
\subsection{Examples of Visual Question Answering (VQA) Types} 

\begin{figure}[htbp]
    \centering
    \includegraphics[width=1.0\linewidth]{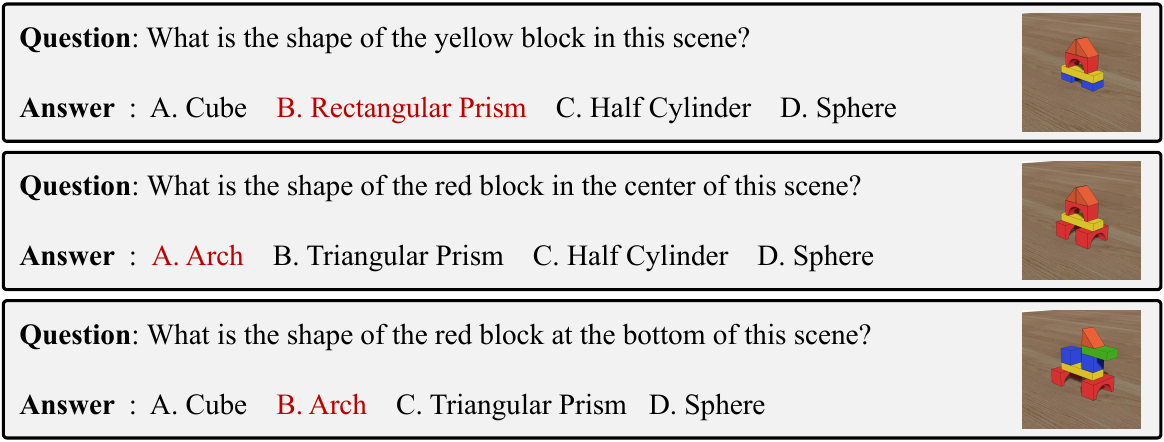}
    \caption{VQA Examples for the \textbf{\textit{Shape}} Subtask}
    \label{fig:01_Shape}
\end{figure}

\begin{figure}[htbp]
    \centering
    \includegraphics[width=1.0\linewidth]{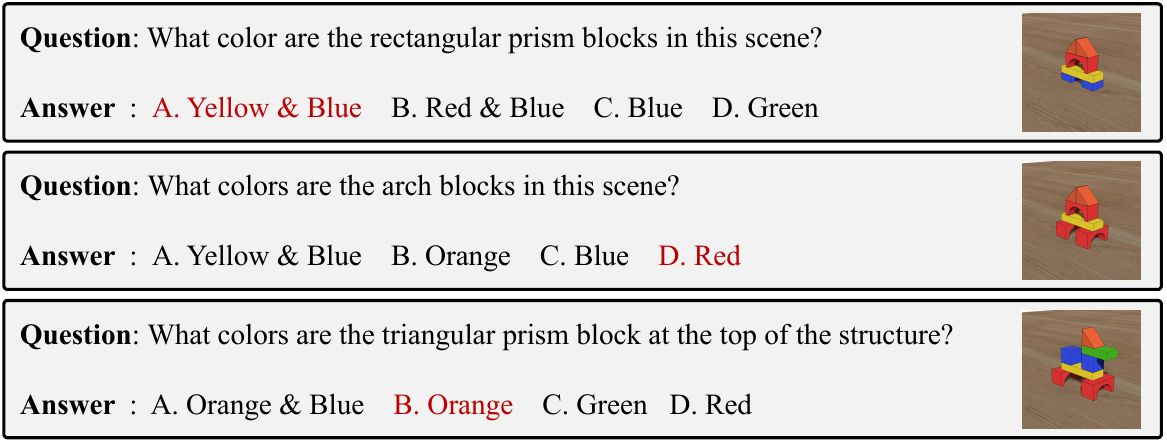}
    \caption{VQA Examples for the \textbf{\textit{Color}} Subtask}
    \label{fig:02_Color}
\end{figure}

\begin{figure}[htbp]
    \centering
    \includegraphics[width=1.0\linewidth]{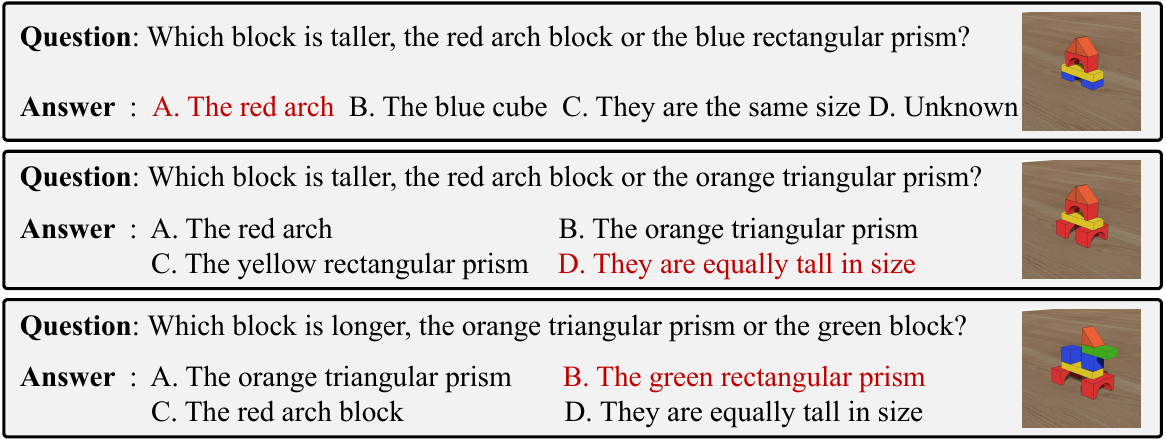}
    \caption{VQA Examples for the \textbf{\textit{Size}} Subtask}
    \label{fig:03_Size}
\end{figure}

\begin{figure}[htbp]
    \centering
    \includegraphics[width=1.0\linewidth]{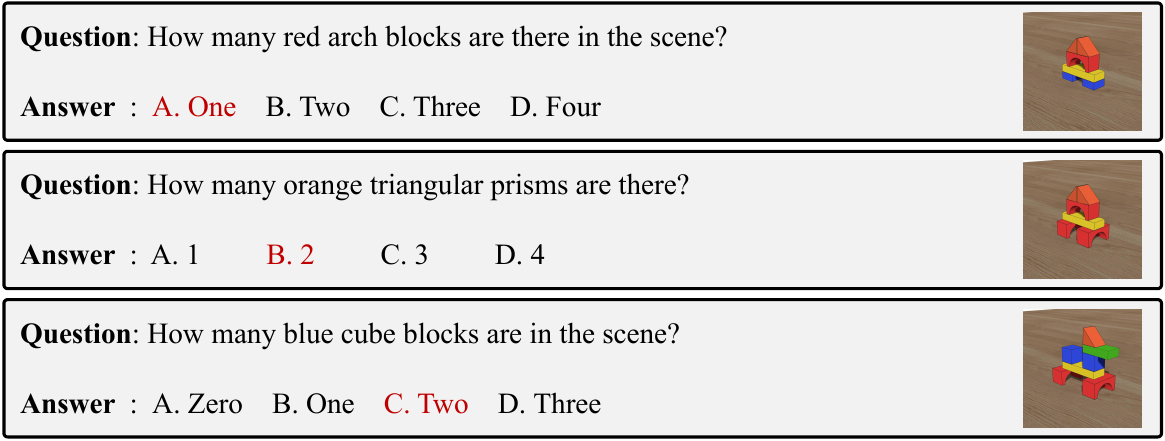}
    \caption{VQA Examples for the \textbf{\textit{Number}} Subtask}
    \label{fig:04_Number}
\end{figure}

\begin{figure}[htbp]
    \centering
    \includegraphics[width=1.0\linewidth]{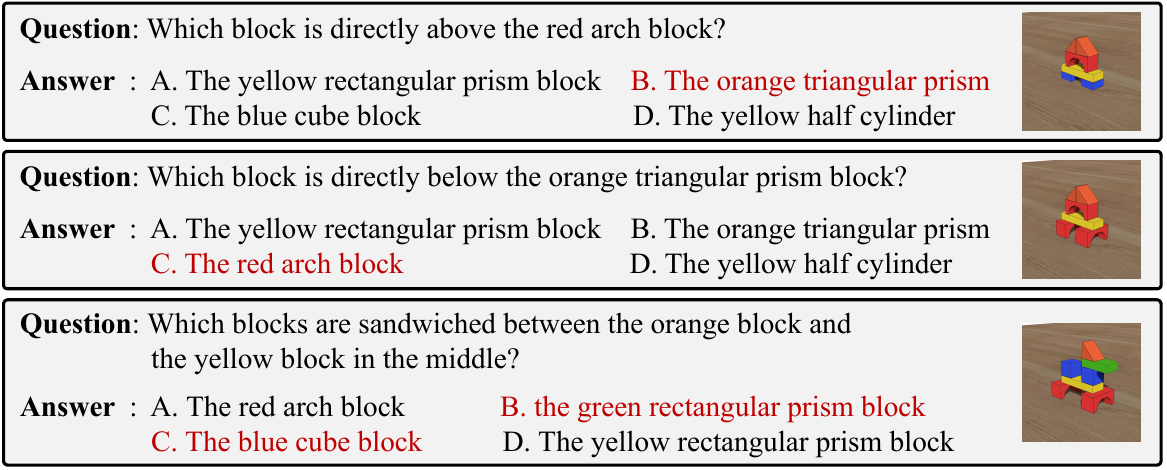}
    \caption{VQA Examples for the \textbf{\textit{Relative Position}} Subtask}
    \label{fig:05_RelativePosition}
\end{figure}

\begin{figure}[htbp]
    \centering
    \includegraphics[width=1.0\linewidth]{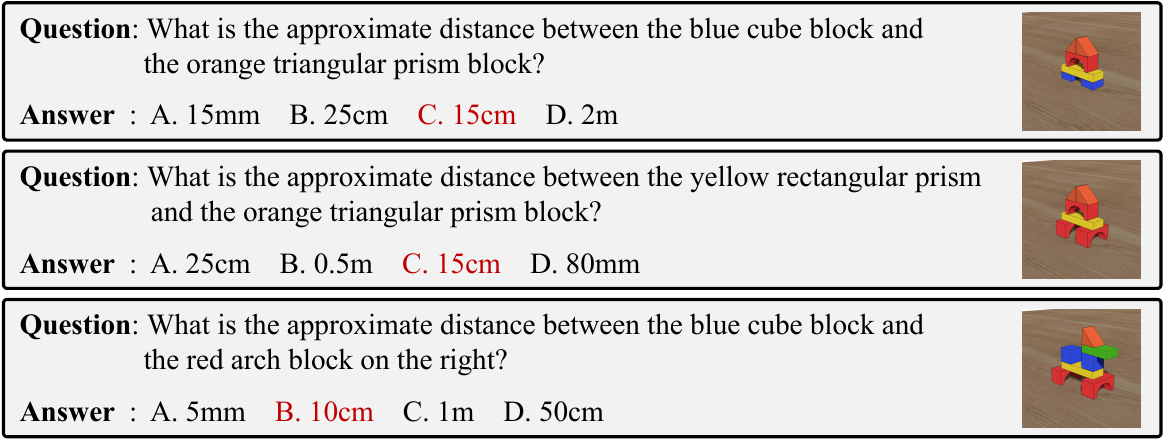}
    \caption{VQA Examples for the \textbf{\textit{Absolute Position}} Subtask}
    \label{fig:06_AbsolutePosition}
\end{figure}

\begin{figure}[htbp]
    \centering
    \includegraphics[width=1.0\linewidth]{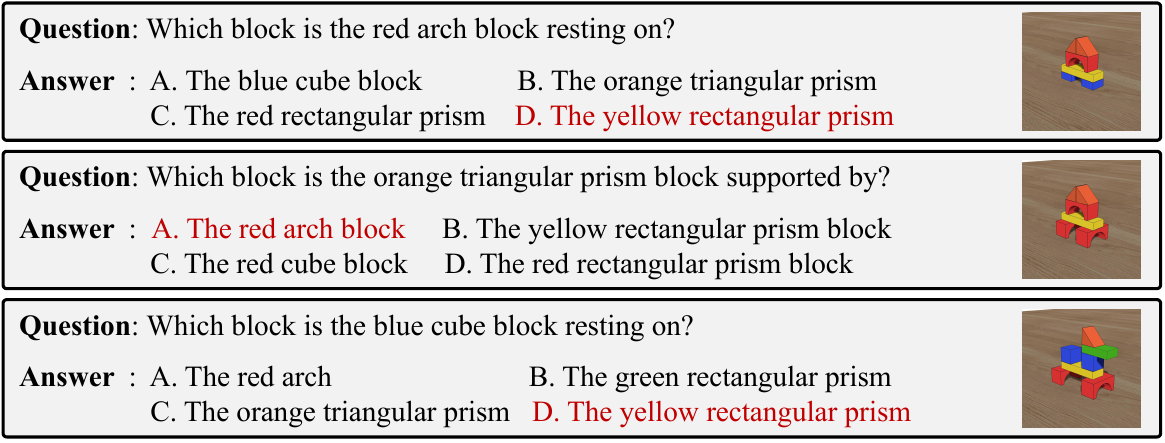}
    \caption{VQA Examples for the \textbf{\textit{Relative Dependency}} Subtask}
    \label{fig:07_RelativeDependency}
\end{figure}

\begin{figure}[htbp]
    \centering
    \includegraphics[width=1.0\linewidth]{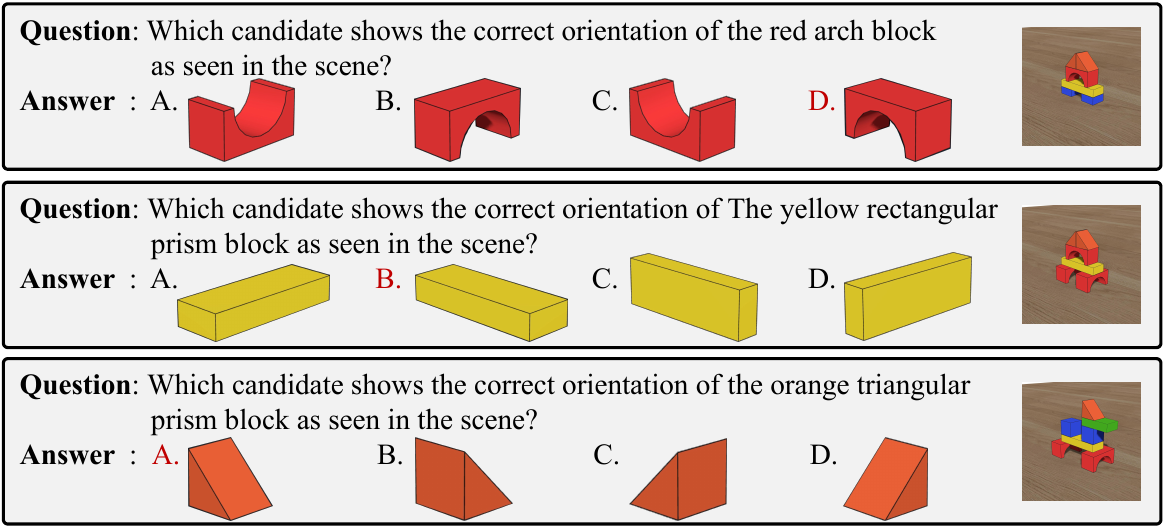}
    \caption{VQA Examples for the \textbf{\textit{Relative Rotation}} Subtask}
    \label{fig:08_RelativeRotation}
\end{figure}

\begin{figure}[htbp]
    \centering
    \includegraphics[width=1.0\linewidth]{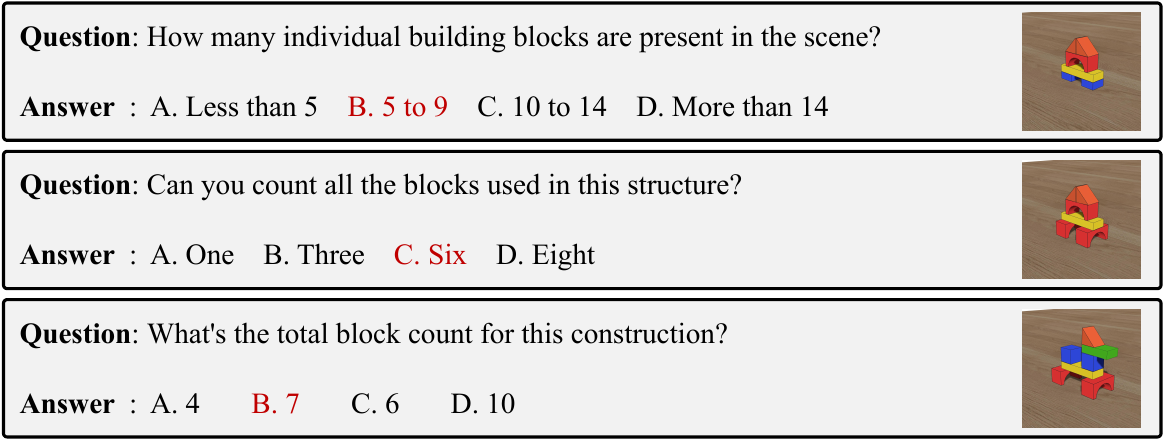}
    \caption{VQA Examples for the \textbf{\textit{Object Counting}} Subtask}
    \label{fig:09_ObjectCounting}
\end{figure}

\begin{figure}[htbp]
    \centering
    \includegraphics[width=1.0\linewidth]{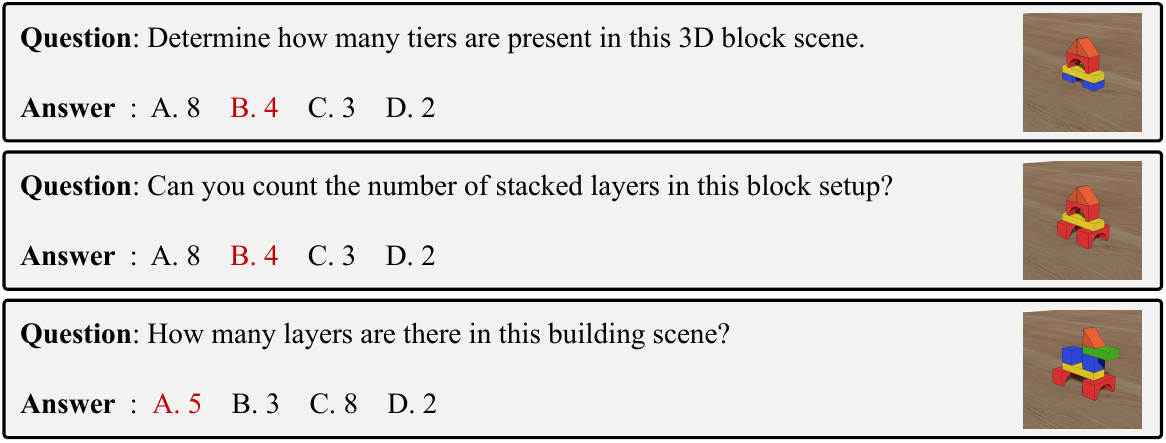}
    \caption{VQA Examples for the \textbf{\textit{Layer Counting}} Subtask}
    \label{fig:10_LayerCounting}
\end{figure}

\begin{figure}[htbp]
    \centering
    \includegraphics[width=1.0\linewidth]{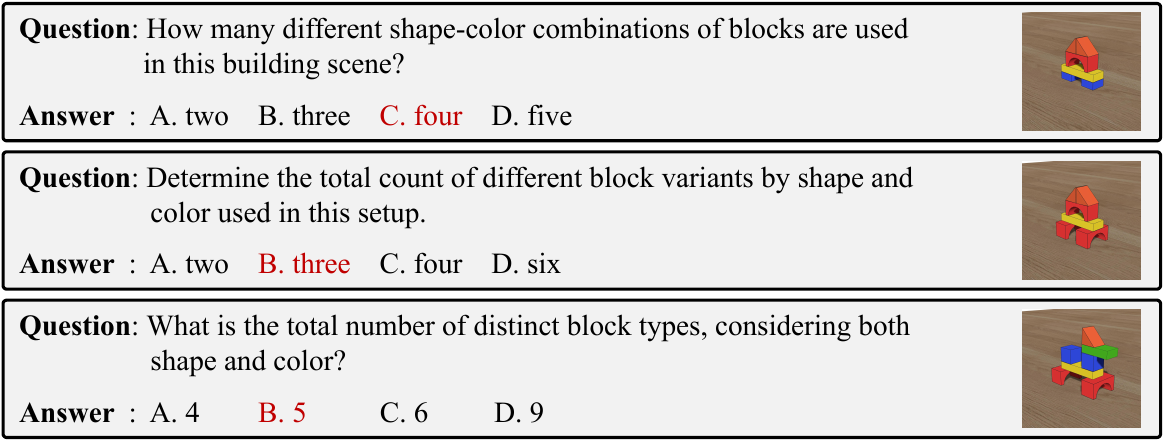}
    \caption{VQA Examples for the \textbf{\textit{Type Counting}} Subtask}
    \label{fig:11_TypeCounting}
\end{figure}

\begin{figure}[htbp]
    \centering
    \includegraphics[width=1.0\linewidth]{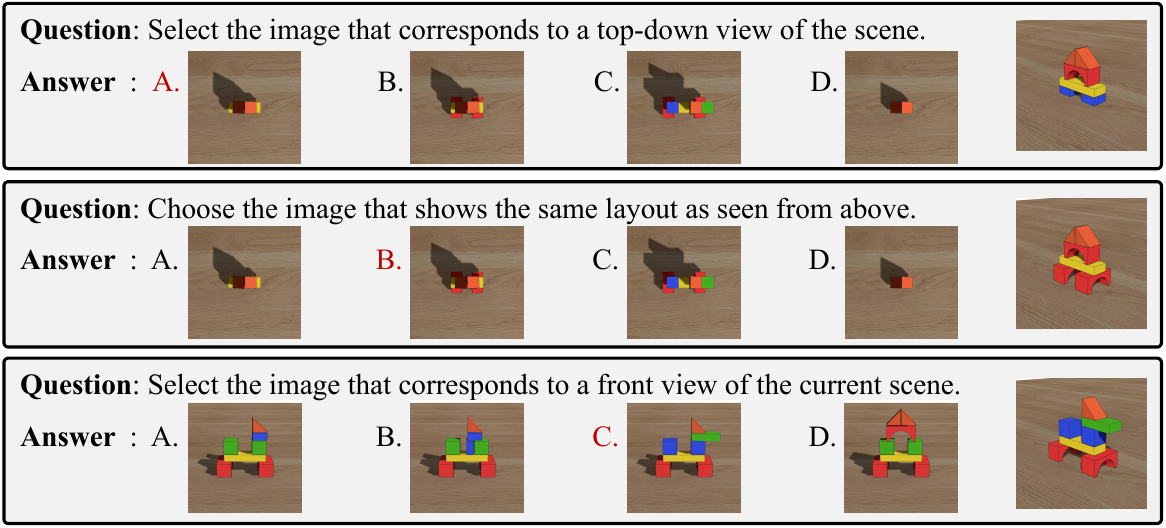}
    \caption{VQA Examples for the \textbf{\textit{Viewpoint}} Subtask}
    \label{fig:12_Viewpoint}
\end{figure}

\begin{figure}[htbp]
    \centering
    \includegraphics[width=1.0\linewidth]{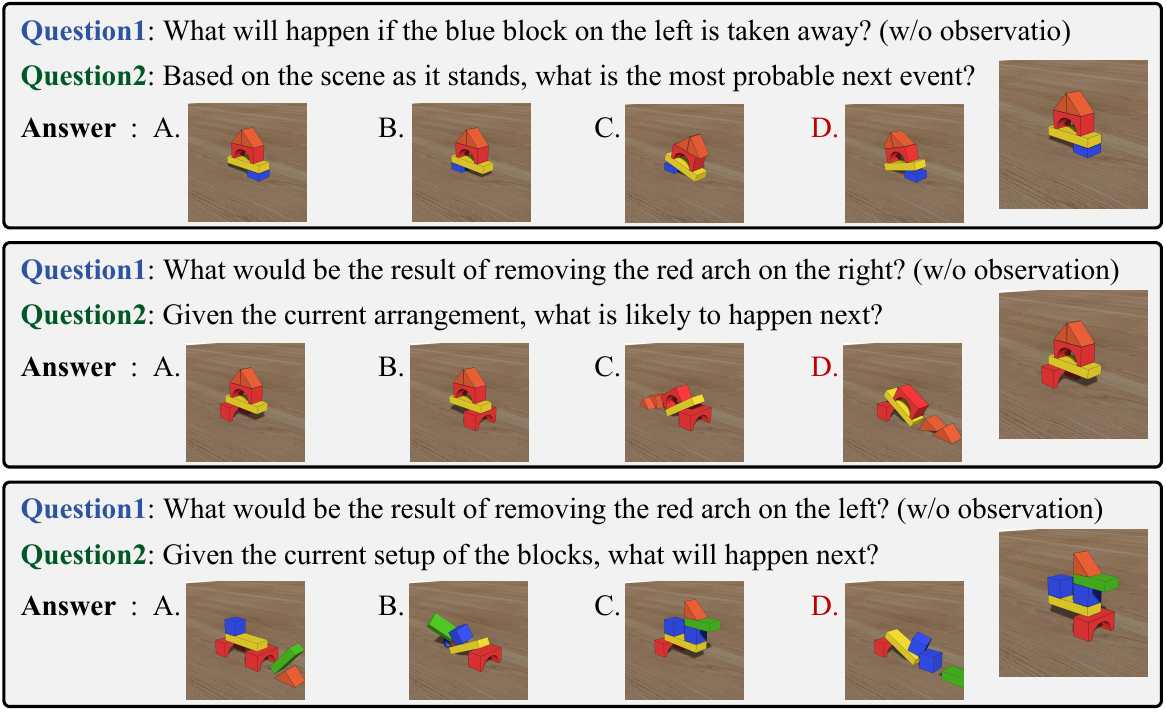}
    \caption{VQA Examples for the \textcolor{myblue}{\textbf{\textit{Counterfactual}}} and  \textcolor{mygreen}{\textbf{\textit{Predictive}}} Subtask}
    \label{fig:13_14_Predict}
\end{figure}

\begin{figure}[htbp]
    \centering
    \includegraphics[width=1.0\linewidth]{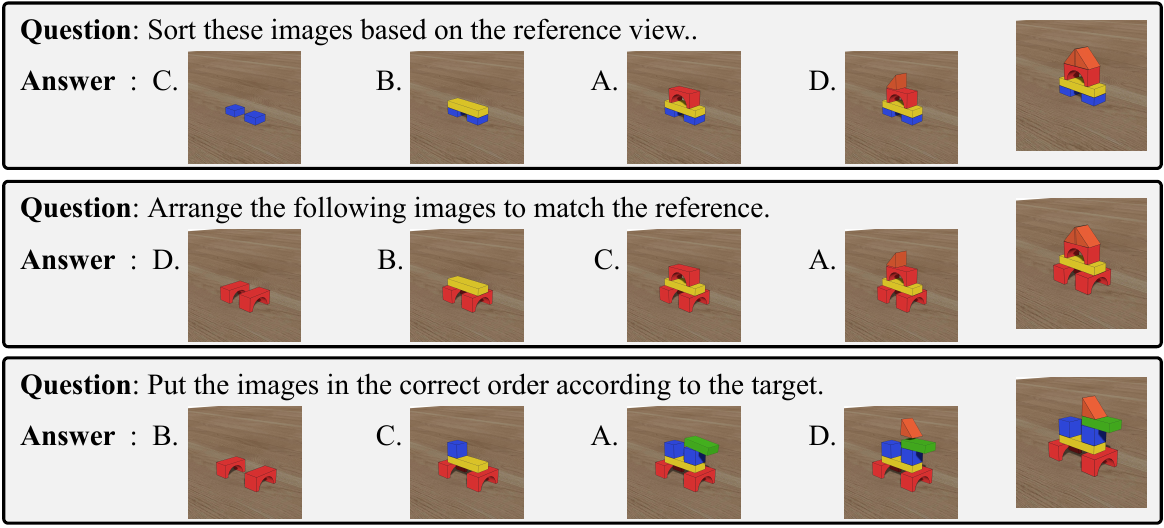}
    \caption{VQA Examples for the \textbf{\textit{Ordering}} Subtask}
    \label{fig:15_Ordering}
\end{figure}

\begin{figure}[htbp]
    \centering
    \includegraphics[width=1.0\linewidth]{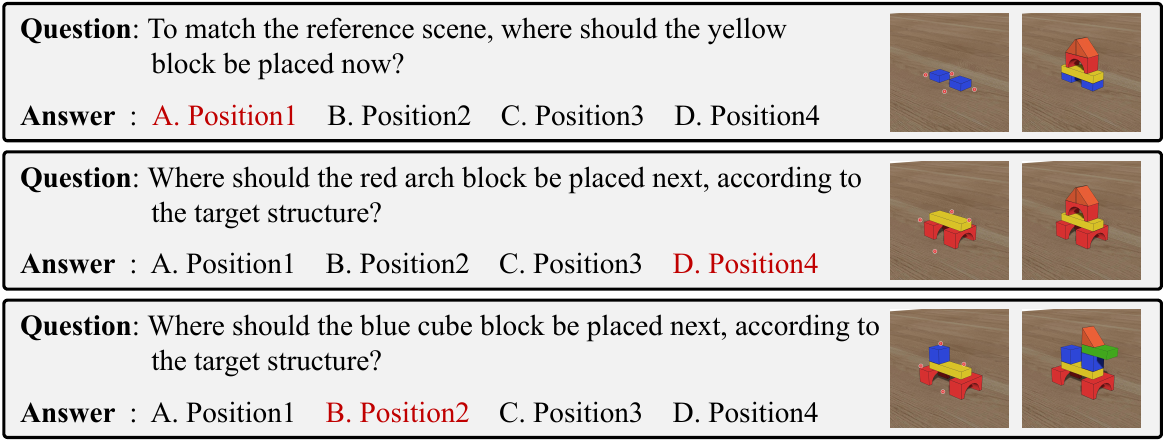}
    \caption{VQA Examples for the \textbf{\textit{Affordance}} Subtask}
    \label{fig:16_Affordance}
\end{figure}

\clearpage
\subsection{Examples of One-time Comprehensive Planning Outputs}

\begin{figure}[htbp]
    \centering
    \includegraphics[width=1.0\linewidth]{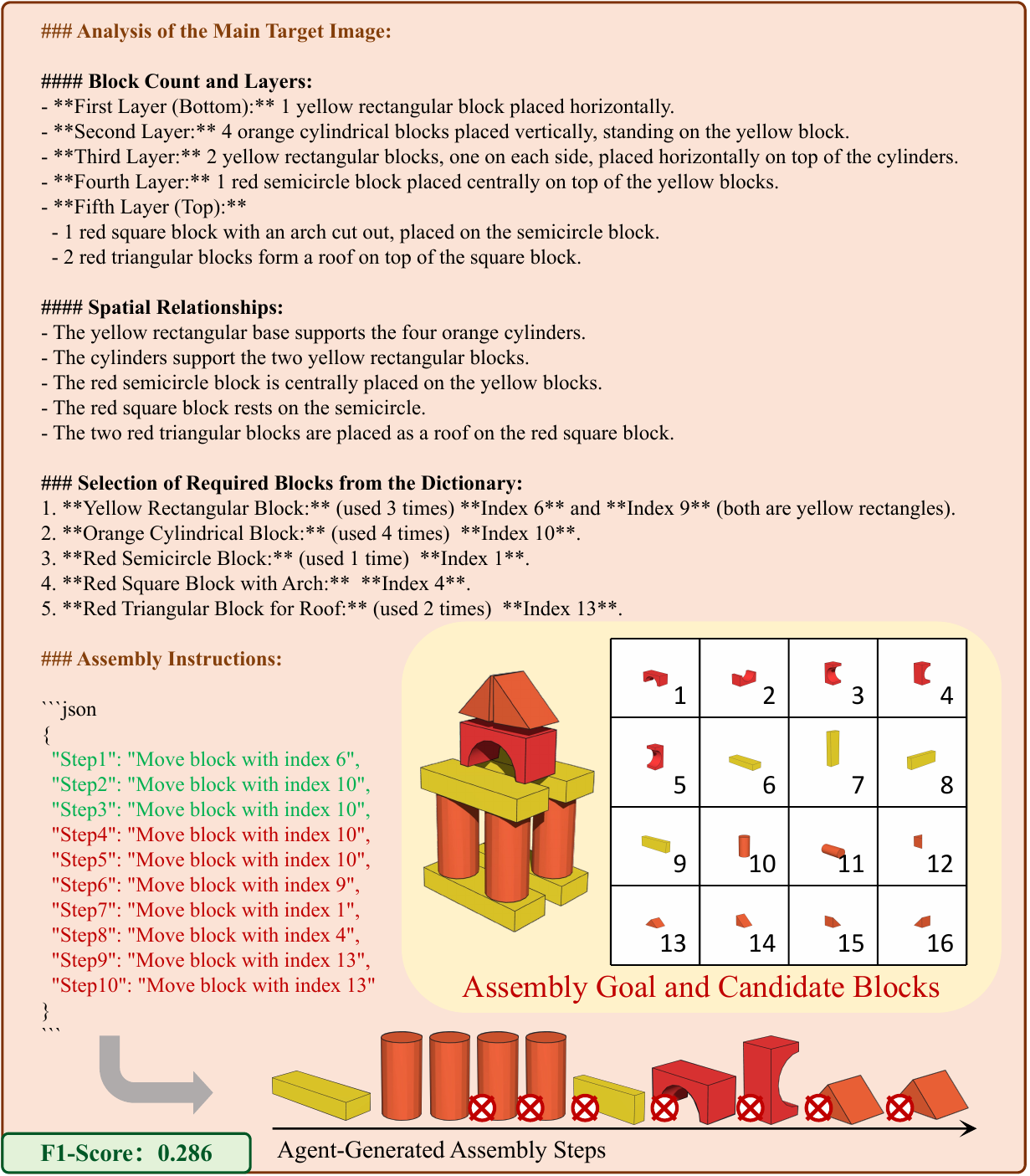}
    \caption{Example inference from \textbf{\textit{GPT-4o}} on scene \textcolor{myblue}{No.029}}
    \label{fig:Output_GPT4o}
\end{figure}


Figure \ref{fig:Output_GPT4o} presents the inference outputs of the GPT-4o on scene No.029. The Assembly Goal and Candidate Blocks illustrate the target assembly configuration image and the set of available building blocks, respectively. Based on the Generated Assembly Steps, we observe that only the first three steps are correct. The two orange cylinders are misclassified as incorrect due to a failure in spatial dependency: they were not placed on top of the yellow cuboid as required. This violation of the dependency constraint led to the collapse of the second layer, causing all subsequent upper-layer blocks to be incorrectly assembled. As a result, the evaluation metrics for this scene are: True Positives (TP) = 3, False Positives (FP) = 7, and False Negatives (FN) = 8, yielding a final F1 score of \textbf{0.286} for the predicted assembly steps.

\clearpage
\begin{figure}[htbp]
    \centering
    \includegraphics[width=1.0\linewidth]{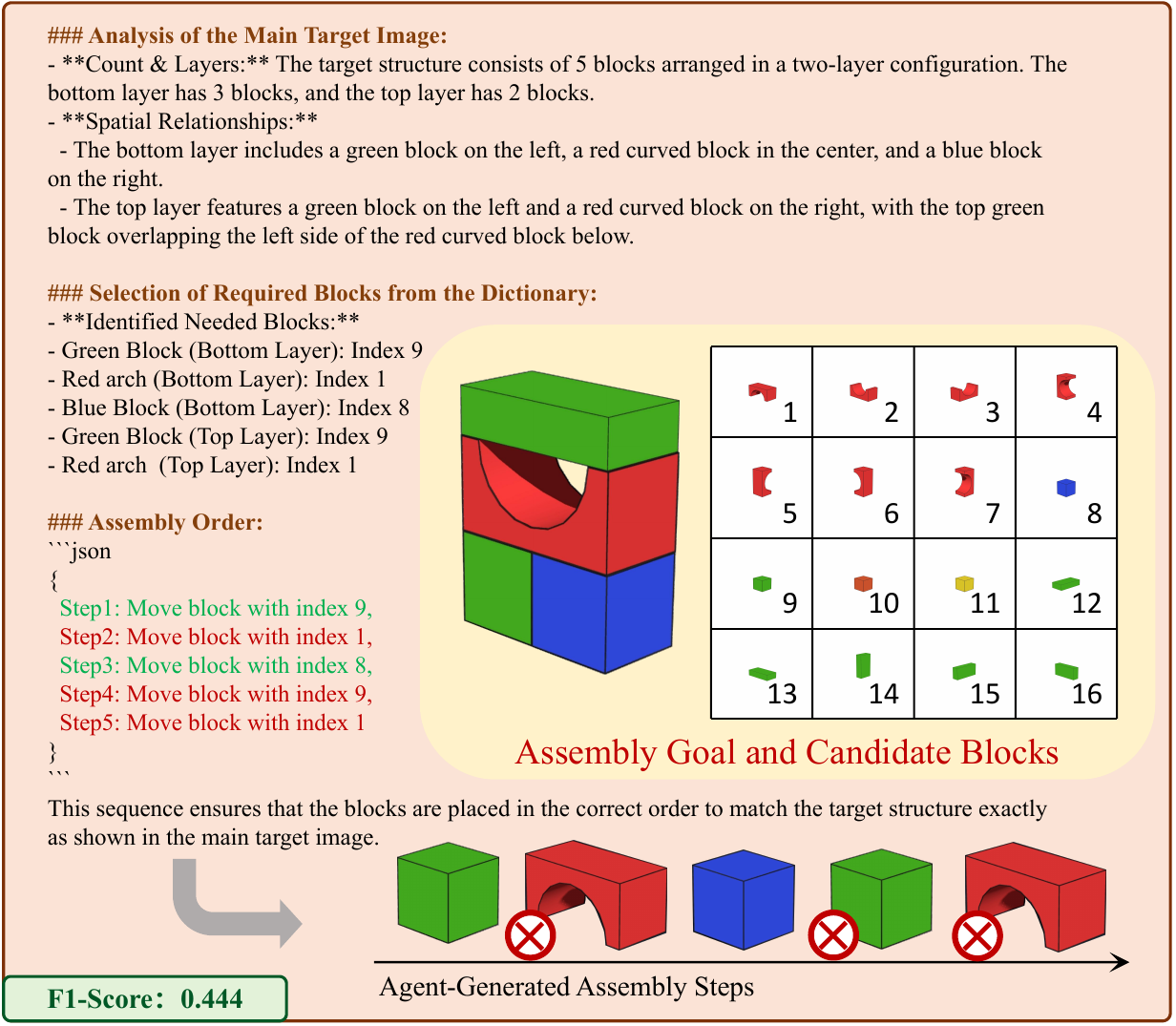}
    \caption{Example inference from \textbf{\textit{Qwen-VL-Max}} on scene \textcolor{myblue}{No.069}}
    \label{fig:Outputs_Qwen}
\end{figure}


Figure \ref{fig:Outputs_Qwen} presents the inference outputs of the Qwen-VL-Max model on scene No.069. The Assembly Goal and Candidate Blocks denote the reference image of the target structure and the set of available building blocks, respectively. From the Generated Assembly Steps, it can be observed that only two steps were executed correctly. Specifically, the red arch block in step two was placed with an incorrect pose, the block selected in step four was a spillover error (i.e., exceeding the necessary components), and the final red arch block was also misaligned due to a pose error. As a result, the evaluation metrics for this scene are: True Positives (TP) = 2, False Positives (FP) = 3, and False Negatives (FN) = 2, yielding a final F1 score of \textbf{0.444} for the predicted assembly trajectory.

\section{More Results and Analysis}
\label{More_results}
\textbf{Repeated evaluations with five random seeds were conducted to ensure robust and reliable model performance.} This approach guarantees that the reported outcomes are not influenced by random fluctuations and provides a more consistent measure of the models' performance. For each model, the mean and standard deviation (±) of the F1 score were computed across various difficulty levels, from Level 1 to Level 4, as well as the overall performance. The evaluation includes both closed-source models, such as GPT-4.1, and open-source models, such as Qwen2.5-VL-7B-Instruct and Cosmos-Reason1-7B. The performance of these models is summarized in Table \ref{tab:model_performance}.

\begin{table}[ht]
    \centering
    \caption{F1 Scores for Model Evaluation across Difficulty Levels.}
    \label{tab:model_performance}
    \setlength\extrarowheight{2pt}
    \normalsize
    \resizebox{\textwidth}{!}{
    \begin{tabular}{l c c c c c}
        \toprule
        \multirow{2}{*}{\bf Model} &  
        \multicolumn{4}{c}{\bf \cellcolor[HTML]{CDD4DF} Difficulty Level} & \textbf{Overall} \\
        
        \cmidrule(lr){2-5} 
        & \textit{Level 1} & \textit{Level 2} & \textit{Level 3} & \textit{Level 4} \\
        
        \midrule
        GPT-4.1 & 94.42 ± 2.3 & 46.26 ± 2.1 & 39.13 ± 2.4 & 36.07 ± 1.3 & 39.73 ± 1.1 \\
        Qwen2.5-VL-7B-Instruct & 44.13 ± 1.3 & 24.12 ± 1.1 & 21.12 ± 2.1 & 16.71 ± 1.7 & 20.40 ± 1.2 \\
        Cosmos-Reason1-7B & 43.45 ± 1.8 & 29.42 ± 2.1 & 21.53 ± 1.7 & 20.42 ± 1.5 & 23.27 ± 1.2 \\
        \bottomrule
    \end{tabular}}
    \vspace{-2mm}
\end{table}

As shown in Table \ref{tab:model_performance}, GPT-4.1 consistently outperforms the open-source models across all levels, with notably low standard deviations (±2.3 to ±1.1), indicating that its performance is stable and reproducible. In contrast, the open-source models exhibit larger variations in their results, suggesting less consistency across different runs. These low error margins for GPT-4.1 confirm the reliability of its performance and demonstrate that the observed differences are not due to random fluctuations. This reinforces the robustness of our evaluation framework and underscores the superior reasoning ability of GPT-4.1, particularly when compared to the open-source models.

\textbf{The comparable F1 scores observed between Level 3 and Level 4 indicate a strong transfer of reasoning patterns across structural complexities.} Upon deeper inspection, our extendable data creation logic involves constructing (partially) Level 4 structures by horizontally merging two different Level 3 structures, while maintaining the same vertical depth. This design allows the reasoning patterns learned for Level 3 to transfer effectively to Level 4, enabling the model to reuse similar inference chains. However, the merging process also introduces spatial reasoning challenges, such as partial visual occlusions and extended reasoning chains, which may compound earlier errors and cause minor inconsistencies. These subtle factors likely explain the observed performance plateau and slight variations across Levels 3 and 4.

\begin{figure}[t]
  \centering
   \includegraphics[width=0.75 \linewidth]{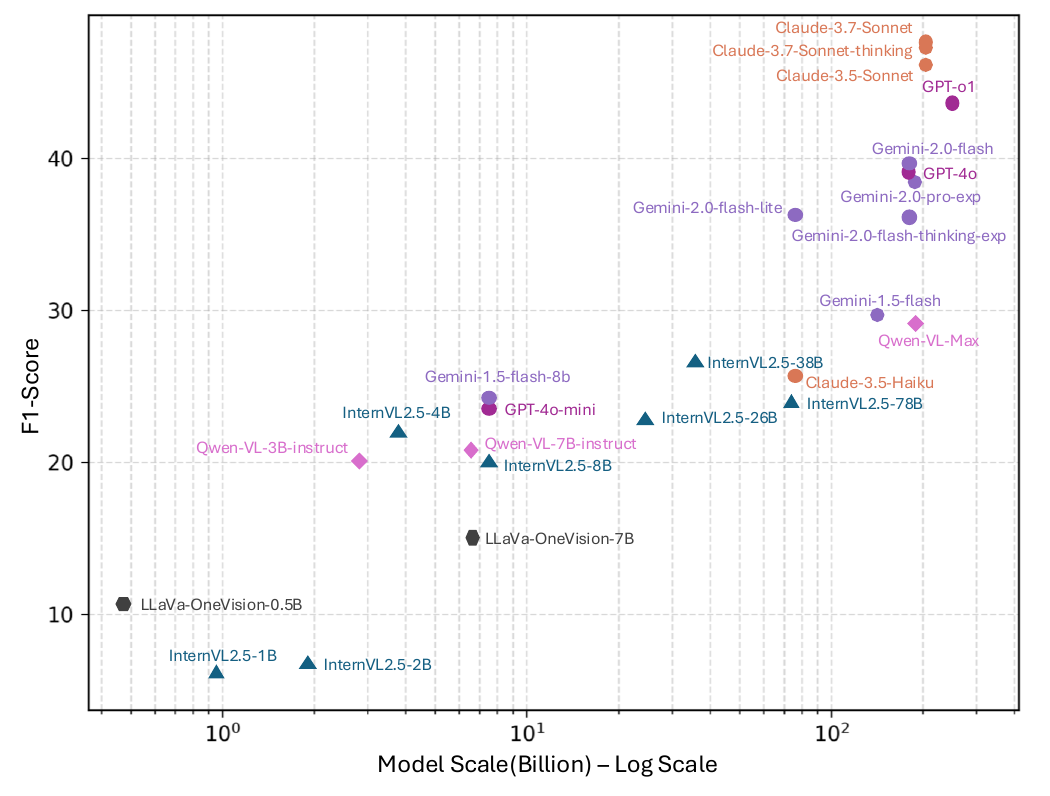}
   \caption{The impact of model size on F\textsubscript{1} Score}
   \label{fig:Fscale}
   \vspace{-4mm}
\end{figure}

\textbf{Robust multi-step reasoning emerges only in >100B-parameter models, whereas smaller models falter.}
As shown in Figure \ref{fig:Fscale}, models with over 100 billion parameters consistently achieve an F\textsubscript{1} score above 30, demonstrating strong reasoning and planning abilities in the block building task. As parameter size decreases, performance declines significantly, with InternVL2.5-1B achieving only F\textsubscript{1} = 5, highlighting the challenges smaller models face in handling multi-step reasoning and spatial constraints. The LeiDA Graph further illustrates performance variations across task difficulty levels, where Claude 3.7 Thinking and GPT-4 maintain relatively strong results across all levels, while smaller models like InternVL2 5-78B and Qwen-VL-Max show inconsistencies, particularly in more complex tasks. These findings emphasize the crucial role of model scale in structured reasoning and multi-step decision-making.

\textbf{Models misjudge block poses, prioritizing color over precise spatial alignment.}
In our experiments as shown in Figure \ref{Fig:Errors}, we found that current vision-language models often exhibit inaccuracies in understanding the spatial poses of building blocks. In our setup, the model must accurately select blocks with spatial poses corresponding to those in the target image to ensure successful assembly. However, existing models tend to prioritize color and block type during selection while neglecting precise spatial alignment. This discrepancy leads to incorrect placements during assembly, compromising structural stability and overall task performance.

\section{Known Limitations and Future Directions of PhyBlock}
\label{Limitations}

\subsection{Limited Inclusion of VLA and Affordance-Centric Models}

While our benchmark comprehensively evaluates 25 of the most powerful vision-language models (VLMs) to date, it does not yet include a systematic assessment of emerging Vision-Language-Action (VLA) models and affordance-centric architectures. This omission is primarily due to the current limitations of these models in performing our 3D block assembly task under a strict zero-shot setting. Nevertheless, we acknowledge the critical importance of these model families in embodied reasoning and real-world interaction. 
At this stage, we prioritize enhancing the reasoning capabilities of VLM-based models on our task, establishing a strong foundation upon which our evaluation and experimentation can be progressively extended to VLA and affordance-centric models.

\subsection{Limitations in 3D Spatial Coverage and Viewpoint Diversity}
To assess the model’s understanding of physical spatial reasoning in 3D block assembly, we design a series of VQA tasks targeting key dimensions such as \textbf{Counting}, \textbf{Rotation}, \textbf{Viewpoint}, \textbf{Ordering}, and \textbf{Affordance}—all closely tied to 3D perception and reasoning. While these tasks aim to comprehensively reflect the model's 3D reasoning and planning capabilities, we acknowledge that our current multi-view setting is limited to four canonical views: front, side, top, and oblique. Although this already distinguishes our benchmark from traditional 2D reasoning tasks, it still falls short for research specifically focused on 3D-awareness. In future work, we plan to incorporate a broader range of viewpoint relationships to better approximate complex and diverse 3D environments.

\subsection{Dataset Scale, Augmentation Potential, and Future Expansion}
Our evaluation dataset comprises 400 distinct 3D block assembly scenarios, from which we construct 2,200 high-quality VQA samples. Notably, thanks to detailed annotations of 150 core scenes—each capturing rich inter-block dependencies—we can readily scale the dataset to millions of configurations through systematic augmentations such as recombination, mirroring, and rotation.

On one hand, we believe that the 400 curated scenarios already cover a broad spectrum of spatial reasoning challenges encountered in 3D block assembly tasks, providing a strong foundation for benchmarking key model capabilities. On the other hand, we have explored more challenging levels involving deeper and denser structures composed of more blocks. 
Our preliminary experiments on such levels using \textbf{\textit{GPT-o1}} reveal a substantial drop in performance, indicating that current models are not yet robust to increased structural complexity.

Therefore, we argue that our current set of 400+ evaluation scenarios is sufficient to probe critical reasoning bottlenecks. Nevertheless, in future iterations of PhyBlock, we plan to extend the scenario set at scale, enabling partitioning into training and evaluation subsets and facilitating the inclusion of fine-tuned models to further advance task-specific performance.

\subsection{Pose Estimation as a Bottleneck for 3D Assembly Tasks}
Our 3D block assembly task is designed to require the model to identify blocks from a candidate set that match the type and orientation of those shown in the target image, and to infer the correct sequence of assembly steps. While this process is relatively straightforward for humans, it remains highly challenging for current vision-language models (VLMs).

A natural question arises: why does our task not require models to predict the exact 3D pose (i.e., position and orientation) of each block in space? In fact, we initially considered this more demanding setting when designing the benchmark. However, through extensive pilot experiments with models such as the GPT and Claude series, we found that current VLMs still struggle significantly with accurate 3D spatial reasoning. Their inability to predict precise poses results in zero completion rates for all block assembly tasks that require pose-level precision, which represents a major performance bottleneck.

Due to this limitation, we opted to simplify the task setting, focusing on type, orientation, and order reasoning while deferring exact pose prediction. Nevertheless, we consider 3D pose estimation a critical frontier and plan to extend our task in future work to include fine-grained pose reasoning.

\subsection{Scope of Evaluation and Sim-to-Real Considerations}
PhyBlock is intentionally designed to evaluate two core competencies of Vision-Language Models: (i) physical and spatial perception, and (ii) high-level assembly planning. Our evaluation setup does not currently include real-robot experiments. Incorporating real-world manipulation would introduce additional challenges, such as grasp synthesis, trajectory optimization, calibration, and hardware reliability. These factors could obscure the benchmark’s diagnostic clarity, making it difficult to determine whether performance limitations stem from reasoning or from actuation. By isolating perception and planning within a physics-accurate simulator, PhyBlock provides precise and reproducible measurements of a model’s reasoning abilities. Nonetheless, we recognize that bridging the sim-to-real gap remains an important future direction. Notably, the action sequences generated in Genesis are already compatible with standard robotic manipulation stacks (e.g., MoveIt and Cartesian impedance controllers), facilitating potential transfer to real-world robotic systems.

\section{Ethics Statement}
Our study focuses on the development and evaluation of a 3D block assembly benchmark (PhyBlock) designed to assess the spatial reasoning and planning capabilities of vision-language models (VLMs). All data used in this benchmark, including rendered scenes and associated VQA questions, were synthetically generated without involving human subjects, sensitive personal data, or real-world environments.

To ensure the transparency and reproducibility of our research, we will make the dataset, benchmark suite, and evaluation protocols publicly available under an appropriate open license. We commit to following best practices in responsible dataset sharing and algorithmic evaluation to support the community in further research while minimizing potential misuse.

We believe this work adheres to the ethical standards outlined by NeurIPS and contributes positively to the development of interpretable and robust multimodal AI systems.

\end{document}

%% file: tables/main_results2.tex
\begin{table*}[t]
\small
    \centering\small
    \caption{Results (\%) overview. Evaluation of 3D Block Assembly Planning (One-time Full Planning).} 
    \label{tab_main:main_results}
    \setlength\tabcolsep{2pt}
    \normalsize
    \resizebox{1.0\textwidth}{!}{
    \begin{tabular}{l c c c c c c c c c c c c c c c}
        \toprule
        \multirow{2.5}{*}{\bf Model} &  
        \multicolumn{3}{c}{\bf \cellcolor[HTML]{F5E6E9} Level-1} & 
        \multicolumn{3}{c}{\bf \cellcolor[HTML]{E2E6E1} Level-2} & 
        \multicolumn{3}{c}{\bf \cellcolor[HTML]{F9F2EB} Level-3} & 
        \multicolumn{3}{c}{\bf \cellcolor[HTML]{D2F4F2} Level-4}  & 
        \multicolumn{3}{c}{\bf \cellcolor[HTML]{CDD4DF} Overall.}\\

        \cmidrule(lr){2-4}\cmidrule(lr){5-7}\cmidrule(lr){8-10}\cmidrule(lr){11-13}\cmidrule(lr){14-16} & 
        \textit{$Rec$} & \textit{$Prec$} & \textit{$F_\textsubscript{1}$} & 
        \textit{$Rec$} & \textit{$Prec$} &  \textit{$F_\textsubscript{1}$} & 
        \textit{$Rec$} & \textit{$Prec$} &  \textit{$F_\textsubscript{1}$} & 
                \textit{$Rec$} & \textit{$Prec$} &  \textit{$F_\textsubscript{1}$} & 
        \textit{$Rec$} & \textit{$Prec$} &  \textit{$F_\textsubscript{1}$} \\
        
        \midrule[0.8pt] 
        Claude-3.5 Haiku               & 59.1 & 41.2 & 48.6 & 39.8 & 27.8 & 32.7 & 31.1 & 20.5 & 24.7 & 28.8   & 16.6 & 21.1 & 32.5 & 20.6 & 25.3 \\
        Claude-3.5 Sonnet                & 74.9 & 72.4 & 73.7 & 58.4 & 56.5 & 57.4 & 44.7 & 41.8 & 43.2 & \textbf{43.9}   & 38.9 & 41.2 & \textbf{47.7} & 44.1 & 45.8 \\
        Claude-3.7 Sonnet                & 75.6 & \textbf{78.0} & \textbf{76.8} & 57.6 & 59.5 & 58.6 & \textbf{46.0} & \textbf{46.0} & \textbf{46.0} & 42.7   & \textbf{40.9} & \textbf{41.8} & 47.6 & \textbf{47.2} & \textbf{47.4} \\
        Claude-3.7 Sonnet-Thinking              & \textbf{75.9} & 77.0 & 76.4 & \textbf{59.6} & \textbf{60.5} & \textbf{60.1} & 45.0 & 44.5 & 44.8 & 42.2   & 40.6 & 41.4 & 47.4 & 46.7 & 47.1 \\
        GPT-4o-mini                      & 55.8 & 42.7 & 48.4 & 34.0 & 26.0 & 29.5 & 28.3 & 20.2 & 23.5 & 25.0   & 15.4 & 19.0 & 28.6 & 19.6 & 23.3 \\
        GPT-4o                           & 69.7 & 67.5 & 68.6 & 50.2 & 48.6 & 49.4 & 39.2 & 36.5 & 37.8 & 35.1   & 31.8 & 33.4 & 40.3 & 37.5 & 38.8 \\
        GPT-o1                               & 69.3 & 72.4 & 70.8 & 50.3 & 52.6 & 51.4 & 41.6 & 42.8 & 42.2 & 39.4   & 39.8 & 39.6 & 42.8 & 43.9 & 43.4 \\
        Gemini-1.5-flash-8b              & 52.0 & 47.0 & 49.4 & 31.4 & 28.4 & 29.8 & 23.1 & 20.9 & 22.0 & 25.4   & 19.6 & 22.1 & 26.0 & 22.1 & 23.9 \\
        Gemini-1.5-flash                 & 62.6 & 54.7 & 58.4 & 38.5 & 33.7 & 35.9 & 30.9 & 25.5 & 28.0 & 31.1   & 22.8 & 26.4 & 33.0 & 26.4 & 29.3 \\
        Gemini-2.0-flash-lite            & 62.8 & 65.3 & 64.0 & 40.3 & 41.9 & 41.1 & 35.6 & 36.4 & 36.0 & 33.6   & 32.8 & 33.2 & 35.9 & 36.3 & 36.1 \\
        Gemini-2.0-flash                 & 68.6 & 66.1 & 67.3 & 46.1 & 44.5 & 45.3 & 40.6 & 37.3 & 38.9 & 38.7   & 33.8 & 36.1 & 41.2 & 37.6 & 39.3 \\

        Gemini-2.0-flash-thinking-exp    & 69.3 & 60.5 & 64.6 & 47.2 & 41.2 & 44.0 & 36.8 & 32.3 & 34.4 & 36.0   & 29.2 & 32.2 & 39.0 & 33.2 & 35.8 \\
        Qwen-VL-Max                      & 61.2 & 48.6 & 54.2 & 40.7 & 32.3 & 36.1 & 33.9 & 26.6 & 29.8 & 29.4   & 19.6 & 23.5 & 34.0 & 25.2 & 28.9 \\
        
        \midrule[0.35pt] 
        
        InternVL2.5-1B                  & 5.3  & 7.3  & 6.2  & 3.0  & 3.8  & 3.3  & 19.3 & 2.5  & 4.4  & 2 & 2.1  & 2.1  & 4.6  & 7.6  & 5.8  \\
        InternVL2.5-8B                  & 44.2 & 39.2 & 41.5 & 24.9 & 22.1 & 23.4 & 22.0 & 16.8 & 19.0 & 23.6   & 13.8 & 17.4 & 23.3 & 16.8 & 19.6 \\
        InternVL2.5-78B                 & 60.1 & 42.1 & 49.5 & 37.6 & 26.3 & 31.0 & 29.2 & 18.4 & 22.6 & 28.3   & 15.1 & 19.7 & 31.2 & 19.0 & 23.6 \\
        Qwen2.5-VL-3B-Instruct           & 37.8 & 36.0 & 36.9 & 25.8 & 24.6 & 25.2 & 20.5 & 15.6 & 17.7 & 25.8   & 13.4 & 17.6 & 23.7 & 16.8 & 19.7 \\
        Qwen2.5-VL-7B-Instruct           & 43.8 & 46.5 & 45.1 & 23.4 & 24.8 & 24.1 & 20.5 & 22.0 & 21.2 & 19.9   & 14.6 & 16.9 & 21.1 & 19.7 & 20.4 \\

        LLaVa-OneVision-0.5B & 43.7 & 16.0 & 23.4 & 29.2 & 10.7 & 15.6 & 19.4 & 3.7  & 6.3  & 24.5   & 6.6  & 10.4 & 24.7 & 6.5  & 10.3 \\
        LLaVa-OneVision-7B   & 38.9 & 21.2 & 27.4 & 24.3 & 13.2 & 17.1 & 19.5 & 12.3 & 15.1 & 22.0   & 9.0  & 12.7 & 21.4 & 11.2 & 14.7\\
         \midrule[0.35pt] 
        Random & 20.4 & 16.5 & 18.3 &  8.6 & 16.7 & 11.4 & 6.7 & 16.1 & 9.5   & 4.1  & 14.4 & 6.4 & 6.1 & 15.8 & 8.8 \\

        \bottomrule
        
    \end{tabular}}
    \vspace{-2mm}
\end{table*}

%% file: tables/vqa_results_new.tex
\captionsetup[table]{skip=15pt}
\begin{table*}[t]
    \centering\small
    \caption{Results (\%) overview. Evaluation of Physical Understanding VQA.} 
    \label{tab_main:main_results_vqa}
    \vspace{-1em}
    \setlength\tabcolsep{2pt}
    \setlength\extrarowheight{2pt}
    \small
    \resizebox{1.0\textwidth}{!}{
    \begin{tabular}{l c c c c c c c c c c c c c c c c c c c c c c}
        \shline
        \multirow{3}{*}{\bf Model} &  
        \multicolumn{5}{c}{\bf \cellcolor[HTML]{F5E6E9} Object Property} & 
        \multicolumn{5}{c}{\bf \cellcolor[HTML]{E2E6E1} Object Relationship} & 
        \multicolumn{5}{c}{\bf \cellcolor[HTML]{F9F2EB} Scene Understanding}  & 
        \multicolumn{5}{c}{\bf \cellcolor[HTML]{D2F4F2} Dynamic Reasoning} &
        \multicolumn{1}{c}{\bf \cellcolor[HTML]{CDD4DF} Overrall.}\\

        \cmidrule(lr){2-6}\cmidrule(lr){7-11}\cmidrule(lr){12-16}\cmidrule(lr){17-21}\cmidrule(lr){22-22} & 
        \textit{$SH$} & \textit{$CO$} & 
        \textit{$SI$} & \textit{$NU$} &
        \textit{$Avg$}&
        \textit{$RP$} & \textit{$AP$} & 
        \textit{$RD$} & \textit{$RR$} & 
         \textit{$Avg$}&
        \textit{$OC$} & \textit{$LC$} & 
        \textit{$TC$} & \textit{$VP$} &
         \textit{$Avg$}&
        \textit{$CF$} & \textit{$PD$} & 
        \textit{$OR$} & \textit{$AD$} &
         \textit{$Avg$}&\textit{$Avg$}
         \\
        
        \shline 
      Claude-3.5 Haiku & 59.3 & 27.3 & 40.7 & 40.0 & 41.8 & 60.0 & 40.0 & 43.3 & 30.0 & 43.3 & 42.0 & 35.3 & 46.7 & 40.7 & 41.2 & 19.3 & 28.0 & 0.0 & 18.0 & 16.3 & 35.7 \\
Claude-3.5 Sonnet & \textbf{92.0} & 77.3 & 51.3 & 20.0 & 60.2 & 76.0 & 65.5 & 88.0 & \textbf{59.3} & \textbf{72.2} & 56.7 & 48.7 & 45.3 & 52.0 & 50.7 & 38.0 & 21.3 & 54.0 & 52.0 & 41.3 & 56.1 \\
Claude-3.7 Sonnet & 91.3 & 77.3 & 52.0 & 44.0 & 66.2 & 79.3 & 59.3 & 85.3 & 47.3 & 67.8 & 52.7 & 54.0 & 53.3 & 35.3 & 48.8 & 23.3 & 14.0 & 50.0 & 40.0 & 31.8 & 54.4 \\
GPT-4o-mini & 52.7 & 31.3 & 46.7 & 58.7 & 47.4 & 58.0 & \textbf{72.0} & 36.0 & 44.7 & 52.7 & 43.3 & 38.0 & 33.3 & 45.3 & 40.0 & 31.3 & 26.0 & 28.0 & 26.0 & 27.8 & 42.0 \\
GPT-4o & 82.0 & 64.7 & 43.3 & 49.3 & 59.8 & 75.3 & 55.3 & 55.3 & 43.3 & 57.3 & 39.3 & 45.3 & 28.7 & 62.7& 44.0 & 38.0 & 36.0 & 34.0 & 28.0 & 34.0 & 48.8 \\
GPT-4.1 & 81.3 & 78.7 & 52.0 & 75.3 & 71.8 & 82.7 & 57.3 & 83.3 & 48.7 & 68.0 & 57.3 & 66.7 & 42.7 & 64.0 & 57.7 & 43.3 & 36.7 & \textbf{74.0} & 60.0 & 53.5 & 62.8 \\
GPT-o3 & 88.0 & \textbf{90.0} & \textbf{70.7} & \textbf{79.3} & \textbf{82.0} & 86.0 & 55.3 & \textbf{86.0} & 54.7 & 70.5 & 41.3 & \textbf{80.0} & 63.3 & \textbf{67.3} & \textbf{63.0} & \textbf{71.3} & \textbf{54.0} & 52.0 & \textbf{80.0} & \textbf{64.3} & \textbf{70.0} \\
Gemini-1.5-flash-8b & 64.0 & 70.7 & 39.3 & 62.7 & 59.2 & 68.7 & 24.7 & 48.7 & 23.0 & 41.3 & 55.3 & 46.0 & 52.7 & 40.7 & 48.7 & 21.3 & 24.7 & 16.0 & 38.0 & 25.0 & 43.6 \\
Gemini-1.5-flash & 86.0 & 78.7 & 48.0 & 71.3 & 71.0 & \textbf{86.7} & 14.0 & 72.7 & 26.0 & 49.9 & \textbf{70.0} & 60.0 & 80.0 & 40.7 & 62.7 & 26.7 & 26.7 & 34.0 & 38.0 & 31.4 & 52.1 \\
Gemini-2.0-flash-lite & 81.3 & 87.3 & 54.7 & 77.3 & 75.2 & 78.0 & 52.0 & 79.3 & 29.3 & 59.7 & 52.0 & 54.7 & 60.7 & 40.7 & 52.0 & 38.7 & 21.3 & 46.0 & 34.0 & 35.0& 55.5 \\
Gemini-2.0-flash & 84.7 & 86.0 & 68.7 & 78.0 & 79.4 & 82.7 & 56.0 & 83.3 & 35.3 & 64.3 & 65.3 & 59.3 & \textbf{81.3} & 40.7 & 61.7 & 39.3 & 24.7 & 70.0 & 38.0 & 43.0 & 62.1 \\
Qwen-VL-Max & 80.0 & 71.3 & 48.7 & 56.7 & 64.2 & 80.0 & 52.0 & 69.3 & 36.0 & 59.3 & 58.7 & 56.0 & 58.0 & 24.7 & 49.4 & 42.7 & 36.7 & 6.0 & 46.0 & 32.9 & 51.6 \\
Qwen2.5-VL-72B-Instruct & 82.7 & 58.0 & 50.7 & 42.0 & 58.4 & 70.7 & 42.7 & 74.3 & 46.0 & 58.4 & 28.0 & 40.0 & 34.0 & 21.3 & 30.8 & 40.7 & 30.0 & 40.0 & 48.0 & 39.7 & 47.1 \\
Random & 24.9 & 25.2 & 25.0 & 24.9 & 25.0 & 23.9 & 23.8 & 23.5 & 23.6 & 23.7 & 25.6 & 25.1 & 25.3 & 24.8 & 25.2 & 27.8 & 26.9 & 4.2 & 27.9 & 21.7 & 24.1 \\

        \shline 
        
    \end{tabular}}
    \caption*{\footnotesize \textbf{Note:} The 16 abbreviations denote different task types, grouped into four categories. 
\textit{SH}: Shape, 
\textit{CO}: Color, 
\textit{SI}: Size, 
\textit{NU}: Number. 
\textit{RP}: Relative position, 
\textit{AP}: Absolute position, 
\textit{RD}: Relative dependency, 
\textit{RR}: Relative rotation. 
\textit{OC}: Object counting, 
\textit{LC}: Layer counting, 
\textit{TC}: Type counting, 
\textit{VP}: Viewpoint.  
\textit{CF}: Counterfactual, 
\textit{PD}: Predictive, 
\textit{OR}: Ordering, 
\textit{AD}: Affordance. }
\vspace{-6mm}
\end{table*}

%% file: tables/block_survey.tex
        




\begin{table}[b]
    \centering\small
    \caption{Comparison of morphological characteristics across different commercial block brands. These features significantly influence the perception and interaction strategies of embodied agents in block-building tasks. "®" denotes registered trademarks of the respective companies. This research is not affiliated with or endorsed by the mentioned companies.} 
    \label{tab:block_survey}
    \setlength\tabcolsep{3pt}
    \setlength\extrarowheight{2pt}
    \normalsize
    \begin{tabular}{l c c c c }
        \shline
        \textbf{Brand} & \textbf{Shape Variety} & \textbf{Color Range} & \textbf{Pattern Types} & \textbf{Connection} \\
        \shline 
        
        LEGO® & 100+ & 12 & Letters / Numbers / Graphics & Slot \\
        Mega Bloks® & 50+ & 7 & Letters / Numbers / Graphics & Slot \\
        MAGFORMERS® & 20+ & 8 & None & Magnetic \\
        BanBao® & 100+ & 11 & None & Slot / Screw \\

        Gigo® & 1 & 10 & None & Slot \\
        Learning Resources® & 4 & 5 & Graphics & Stack \\
        TopBright® & 6 & 10 & Letters / Numbers / Graphics & Stack \\
        Hape® & 8 & 8 & Letters / Numbers / Graphics & Stack \\

        MuWanShiJia® & 9 & 8 & Letters / Numbers / Graphics & Stack \\
        LeLeFish® & 1 & 5 & None & Stack \\

        \shline 
    \end{tabular}
\end{table}

%% file: Figs/Appendix_A/AssemPlan_Blocks_MaxAvg.tex
\begin{figure}[t]
    \centering
    \begin{subfigure}{6.5cm}
        \centering
        \includegraphics[width=1.0\linewidth]{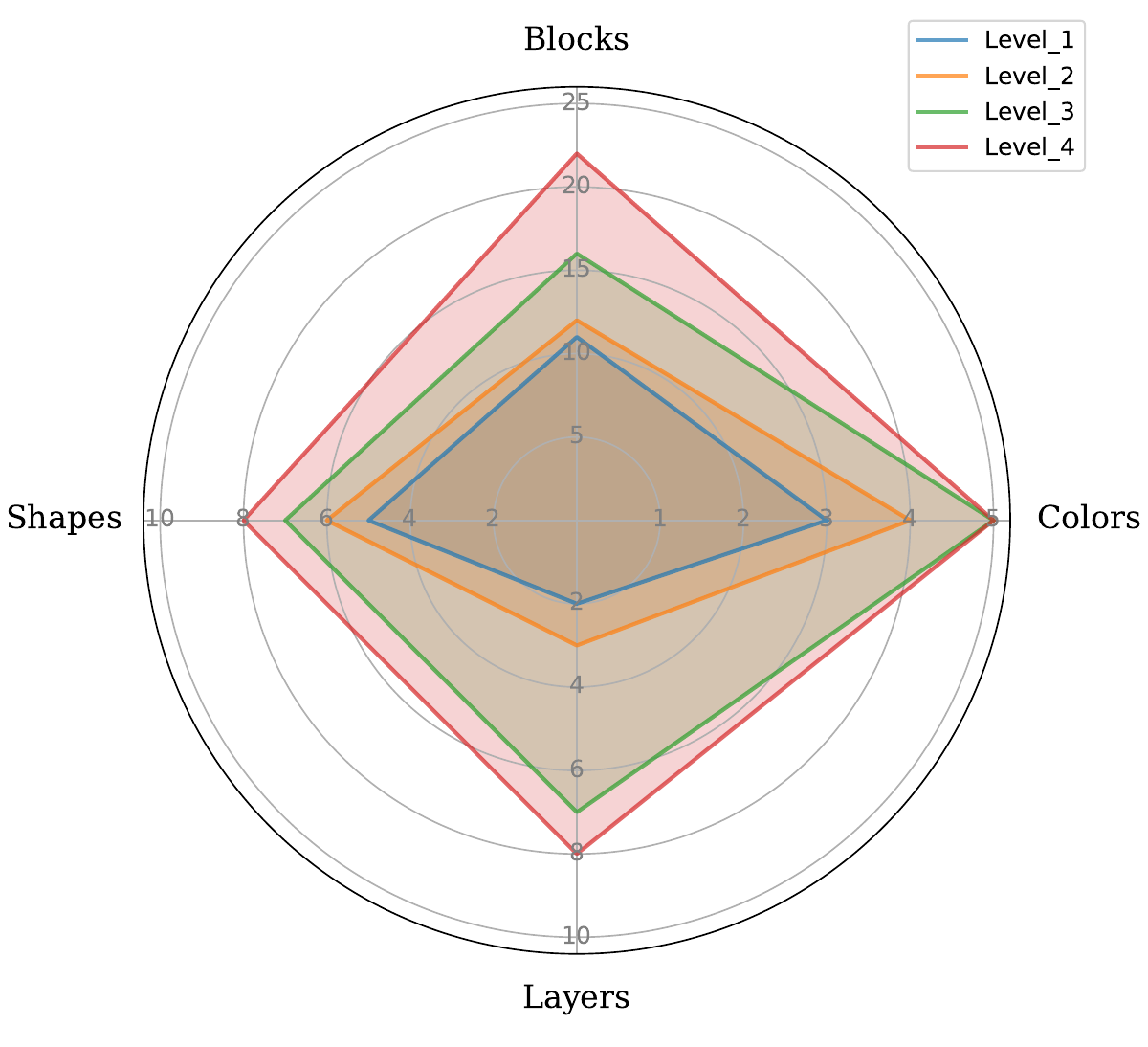}
        \caption{Maximum Value Distribution}
    \end{subfigure}
    \hspace{4pt}
    \begin{subfigure}{6.5cm}
        \centering
        \includegraphics[width=1.0\linewidth]{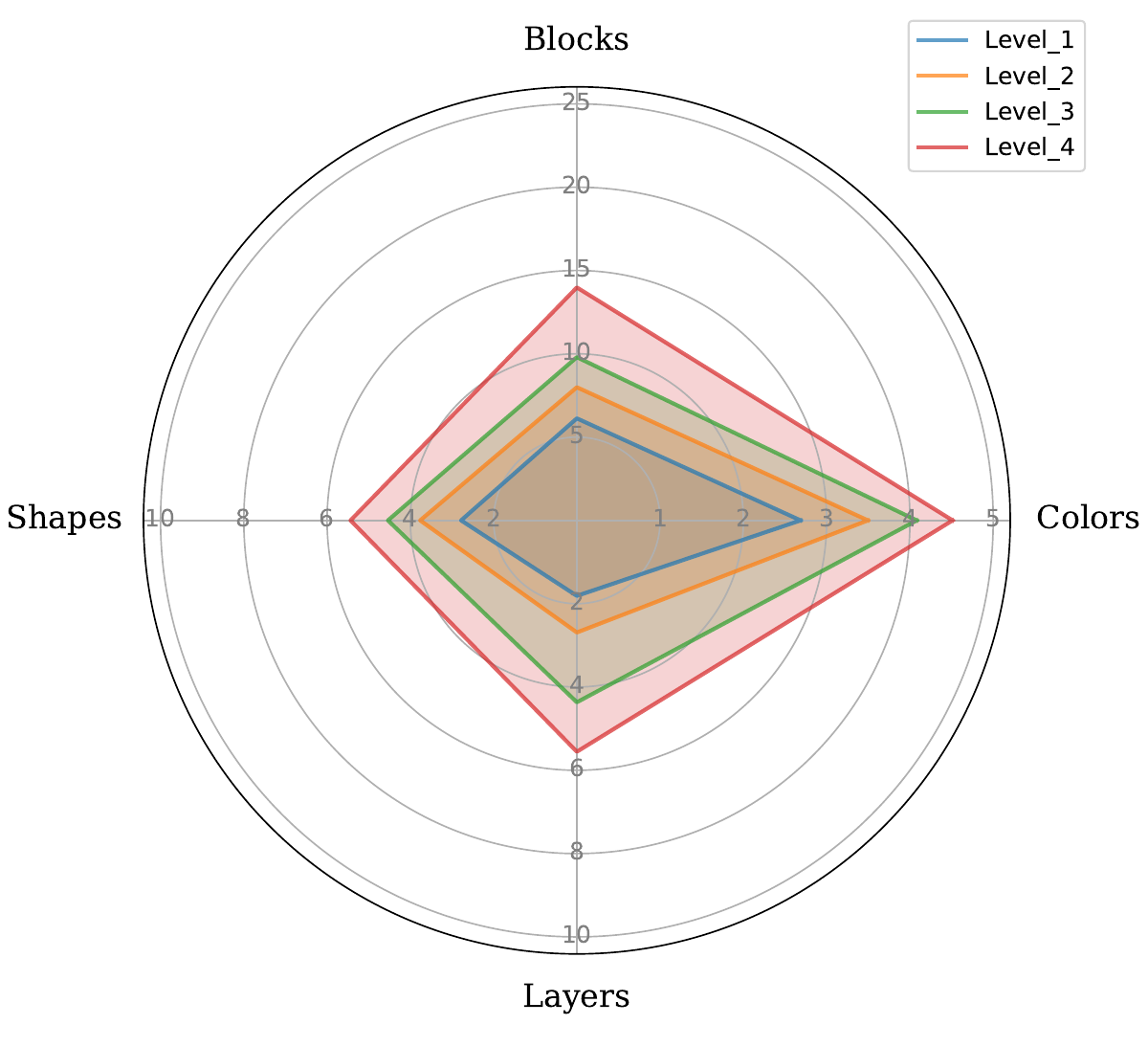}
        \caption{Average Value Distribution}
    \end{subfigure}
    \caption{Distribution of Block Assembly Scenes Across Four Evaluation Levels}
    \label{fig:maxavg}
    \vspace{-12pt}
\end{figure}

%% file: tables/AOV_algo.tex
{\renewcommand{\baselinestretch}{1.2}\selectfont
\begin{algorithm}[t]
\KwIn{
    Ground truth assembly sequence of blocks $GT$,\\
    \hspace{3.3em}Predicted assembly sequence of blocks $P$
}
\KwOut{
    Evaluation metrics: True Positives ($TP$), False Positives ($FP$), False Negatives ($FN$)
}

\tcc{Step 1: Initialization}
\ForEach{$block_{GT} \in GT$}{
    $block_{GT}.placed \leftarrow \text{False}$\;
}
\ForEach{$block_{P} \in P$}{
    $block_{P}.matched\_order \leftarrow 0$\;
}

\tcc{Step 2: Match each predicted block to the earliest valid GT block}
\ForEach{$block_{P} \in P$}{
    \ForEach{$block_{GT} \in GT$}{
        \If{
            $block_{GT}.\text{placed} = \text{False}$ \textbf{and}\\
            \hspace{1.5em}$block_{GT}.\text{is\_place\_legal}$ \textbf{and}\\
            \hspace{1.5em}$block_{P}.type = block_{GT}.type$ \textbf{and}\\
            \hspace{1.5em}$block_{P}.pose = block_{GT}.pose$
        }{
            $block_{GT}.placed \leftarrow \text{True}$\;
            $block_{P}.matched\_order \leftarrow block_{GT}.order$\;
            \textbf{break}\;
        }
    }
}

\tcc{Step 3: Count evaluation metrics}
$TP \leftarrow$ number of $block_{P}$ where $matched\_order \neq 0$\;

$FP \leftarrow$ number of $block_{P}$ where $matched\_order = 0$\;

$FN \leftarrow |GT| - TP$\;

\KwRet{$TP$, $FP$, $FN$}
\caption{Evaluation Algorithm for Predicted Block Assembly Sequences}
\label{algo:block_eval}
\end{algorithm}
}